\documentclass{article}

\usepackage{iclr2025_conference}

\usepackage[utf8]{inputenc} 
\usepackage[T1]{fontenc}    
\usepackage{hyperref}       
\usepackage{url}            
\usepackage{booktabs}       
\usepackage{amsfonts}       
\usepackage{nicefrac}       
\usepackage{microtype}      
\usepackage{xcolor}         
\hypersetup{
    colorlinks=true,    
    linkcolor=blue,     
    citecolor=blue,     
    filecolor=blue,     
    urlcolor=blue       
}
\usepackage{times} 
\usepackage{latexsym} 
\usepackage{amsmath} 
\usepackage{amssymb} 
\usepackage{mathtools} 
\usepackage{annotate-equations} 
\usepackage{pifont} 
\usepackage{truncate} 
\usepackage{enumitem} 
\usepackage{csquotes} 
\usepackage{xurl} 
\usepackage[dont-mess-around]{fnpct} 
\usepackage{soul} 
\usepackage{listings}
\usepackage{algorithm}
\usepackage{algpseudocode}
\algrenewcommand\algorithmicrequire{\textbf{Input:}}
\algrenewcommand\algorithmicensure{\textbf{Output:}}
\newcommand{\CommentTriangle}[1]{%
  \hfill\(\triangleright\) {\scriptsize#1}%
}

\MakeRobust{\Call}

\usepackage{ragged2e} 
\usepackage{float} 

\usepackage{caption}

\usepackage{graphicx} 
\usepackage{svg}

\usepackage{array}
\usepackage{booktabs}
\usepackage{multirow}
\usepackage{tabularx}
\usepackage{tabulary}
\usepackage{tabto}
\usepackage{longtable}
\usepackage{float}

\usepackage{tikz,pgfplots}
\pgfplotsset{compat=1.17}

\usepackage{xinttools} 
\usepackage{ifthen} 

\usepackage{lipsum}

\usepackage{todonotes}

\usepackage[T1]{fontenc}

\newcommand{\cmmnt}[1]{}

\newcommand{\mainfig}{
    \begin{figure}[t]
        \centering
        \includegraphics[width=0.95\linewidth]{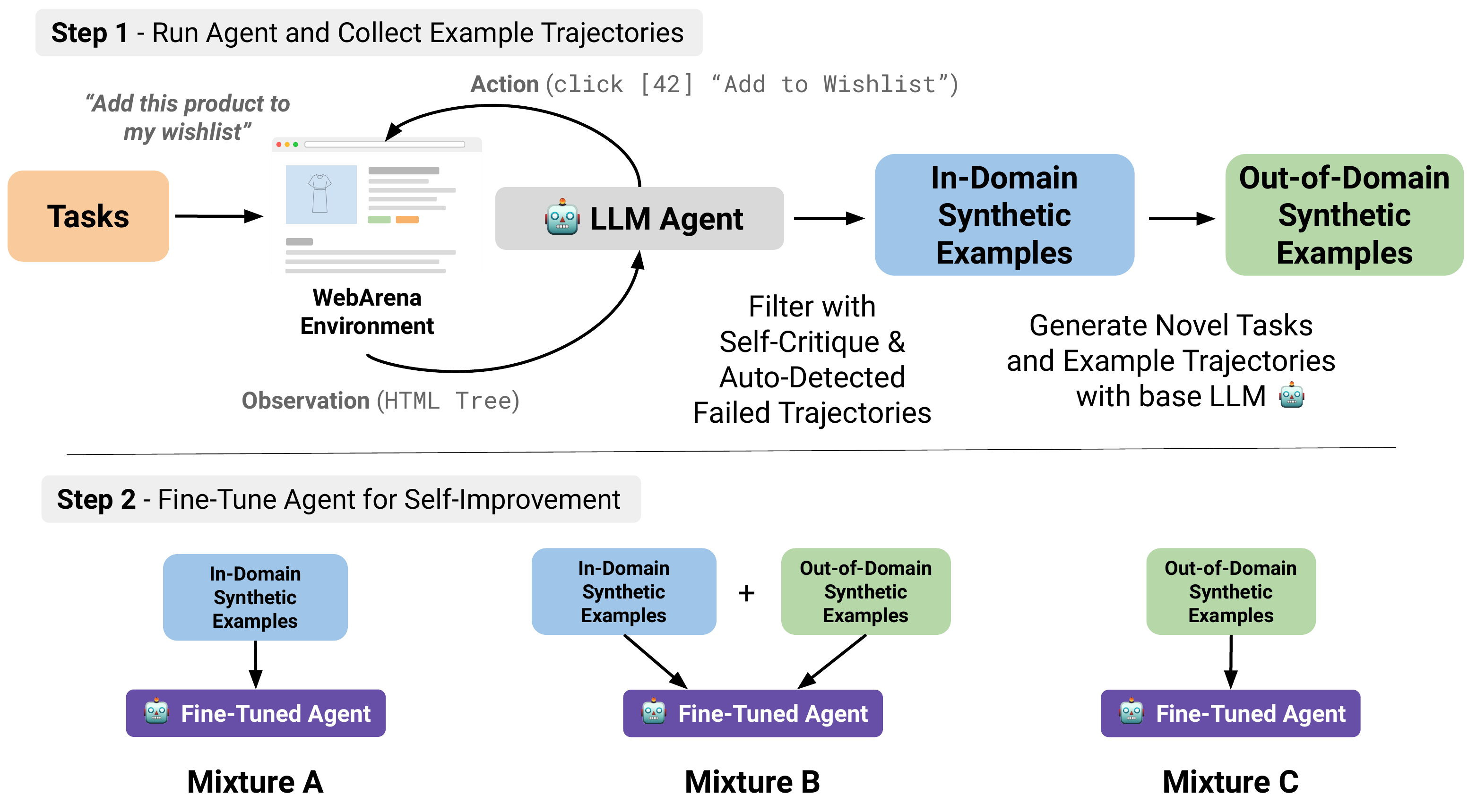}
        \caption{We generate synthetic data to fine-tune LLM agents to accomplish WebArena tasks such as ``{\it Add this product to my wishlist}''. \textbf{Step 1:} We first collect an initial set of trajectories, filter out low-quality trajectories in an unsupervised fashion, and keep the remainder as synthetic in-domain examples. We prompt our base LLM to generate novel out-of-domain tasks along with hypothetical solution trajectories by providing a few in-domain examples. \textbf{Step 2:} We then fine-tune our base LLM agent on each of the three distinct synthetic training data mixtures and evaluate performance.}
        \label{fig:main}
    \end{figure}
}

\newcommand{\hyperparamtable}{
    \begin{table}[H]
    \small
    \centering
    \begin{tabular}{m{.62\textwidth}>{\centering\arraybackslash}m{.3\textwidth}}
    \toprule[\heavyrulewidth]
    \textbf{Hyperparameter} & \textbf{Value} \\ \midrule
    Model & Qwen/Qwen1.5-72B-Chat \\
    Hardware & 2x NVIDIA RTX A6000 \\
    Distributed Protocol & PyTorch FSDP \\
    Data Type & \small{\texttt{torch.bfloat16}} \\
    Quantization & 4-bit (nf4), double quantized \\
    LoRA & \small{\texttt{all-linear, r=8\newline \hspace{4em} lora\_alpha=8 \newline lora\_dropout=0.0}} \\
    Optimizer & \texttt{adamw\_torch} \\
    Learning Rate & 1e-5 \\
    Weight Decay & 0.01 \\
    Learning Rate Scheduler & \small{\texttt{linear}} \\
    Warmup Steps & 0 \\
    Batch Size & 16 \\
    Train-Validation Split & 90/10\% \\
    Early Stopping Threshold & 0.0 \\
    Early Stopping Patience & 5 epochs \\

    \bottomrule[\heavyrulewidth]
    \end{tabular}
    \caption{Hyperparameters selected for fine-tuning experiments.}
    \label{table:hyperparamtable}
    \end{table}
}

\newcommand{\inferenceparamtable}{
    \begin{table}[H]
    \small
    \centering
    \begin{tabular}{m{.62\textwidth}>{\centering\arraybackslash}m{.3\textwidth}}
    \toprule[\heavyrulewidth]
    \textbf{Inference Parameter} & \textbf{Value} \\ \midrule
    Model & Qwen/Qwen1.5-72B-Chat \\
    Hardware & 4x NVIDIA RTX A6000 \\
    Data Type & \small{\texttt{torch.bfloat16}} \\
    Quantization & 4-bit (nf4), double quantized \\
    Prompt Template & \texttt{\small{p\_cot\_id\_actree\_2s}} \\
    Temperature & 1.0 \\
    Top-P & 0.9 \\
    Max New Tokens & 384 \\

    \bottomrule[\heavyrulewidth]
    \end{tabular}
    \caption{Parameters used during inference, we follow the default parameters for inferencing set by the WebArena benchmark \citep{webarena}.}
    \label{table:inferenceparamtable}
    \end{table}
}
\newcommand{\indomainfilteringtable}{
    \begin{table}[H]
    \small
    \centering
    \begin{tabular}{lrrrrr}
    \toprule[\heavyrulewidth]
    \textbf{Set of Trajectories} & \textbf{\#} & \textbf{Accuracy} & \textbf{F1} & \textbf{Precision}  & \textbf{Recall} \\ \midrule
    All Trajectories & 812 & 0.071 & 0.133 & 0.071 & 1.000 \\
    Plausible Trajectories $\mathcal{P}$ & 58 & 0.919 & 0.431 & 0.431 & 0.431  \\

    \bottomrule[\heavyrulewidth]
    \end{tabular}
    \caption{Metrics on the proportion of trajectories that successfully completed the task in the set of plausible trajectories kept in $\mathcal{P}$ after filtering out low-quality trajectories. Approximately 43\% of trajectories in $\mathcal{P}$ successfully completed the task, up from \textasciitilde 7\% with no filtering, indicating useful learning signal is introduced by filtering using self-critiques and information from the environment.}
    \label{table:indomainfiltering}
    \end{table}
}
\newcommand{\outofdomainexamplestable}{
    \begin{table}[H]
{\small

\begin{tabularx}{\textwidth}{>{\centering\arraybackslash}X>{\centering\arraybackslash}X}
\toprule[\heavyrulewidth]
\textbf{Objectives in WebArena Benchmark} & \textbf{Generated Out-of-Domain Objectives} \\
\hline
{\flushleft
\begin{itemize}[noitemsep]
    \item Tell me the total cost of my latest pending order? (Shopping)
    \item Compare the time for walking and driving route from AMC Waterfront to Univ of Pittsburgh (Maps)
    \item Check out the most recent open issues (GitLab)
    \item Which customer has placed 2 orders in the entire history? (Shopping Admin)
    \item $\ldots$
\end{itemize}
}
& 
{\flushleft
\begin{itemize}[noitemsep]
    \item Locate and purchase a subscription to The Economist digital edition (\texttt{\tiny https://store.economist.com/...})
    \item Find the nutrition facts for a Grilled Chicken Caesar Salad from Chili's (\texttt{\tiny http://www.chilis.com/...})
    \item Find the active coupons for a one-year subscription to Adobe Creative Cloud (\texttt{\tiny https://www.couponcabin.com/...})
    \item Subscribe to the premium plan for Grammarly to unlock advanced writing features. (\texttt{\tiny https://www.grammarly.com/...})
    \item $\ldots$
\end{itemize}
}
\\
\bottomrule[\heavyrulewidth]
\end{tabularx}
}
    \caption{A sample of the novel objectives generated compared with the objectives found in WebArena. A full sample of a generated out-of-domain synthetic example can be found in Appendix \ref{sec:full_outofdomain_example}. }
    \label{table:outofdomainexamples}
    \end{table}
}
\newcommand{\fulloutofdomainexampletable}{
    \begin{table}[h!]
\renewcommand{\arraystretch}{1.5}
{\small
\begin{tabularx}{\textwidth}{m{.25\textwidth}m{.68\textwidth}}
\toprule[\heavyrulewidth]
\textbf{Objective} & Subscribe to the premium plan for Grammarly to unlock advanced writing features. \\
\textbf{URL} & \texttt{\tiny https://www.grammarly.com/plans} \\
\textbf{Plan} & 
{
\tiny{
1. Click on the ``Go Premium'' button located at the top right corner of the page.
\newline 2. Scroll down to review the available plans (if needed).
\newline 3. Click on the ``Choose Plan'' button for the preferred premium plan.
\newline 4. Provide payment information, such as credit card details, in the respective input fields.
\newline 5. Enter your billing information, including name, address, and zip code (if required).
\newline 6. Type your email address again to confirm it.
\newline 7. Create a password for your Grammarly account and type it into the password field.
\newline 8. Check the box to agree to the terms and conditions.
\newline 9. Click on the ``Start Your Free Trial'' (or similar) button if you qualify for a trial.
\newline 10. Review the confirmation message and wait for the subscription activation.
\newline 11. Stop when the subscription is successfully activated and the premium features are unlocked.
}} \\
\textbf{Selected Step in Plan} & 3 \\
\textbf{Previous Action} & click [1098] where [1098] is Generic \textquotesingle Premium\textquotesingle \\
\textbf{Next Action} & Let's think step-by-step. To proceed with subscribing to the premium plan, the user must now choose their plan. They have already clicked on the \textquotesingle Premium\textquotesingle ~plan option. The next logical action is to click on the \textquotesingle Choose Plan\textquotesingle ~button to finalize the selection and proceed to the payment page. In summary, the next action I will perform is \textasciigrave\textasciigrave\textasciigrave click [1105]\textasciigrave\textasciigrave\textasciigrave ~to select the premium plan and move to the next step in the subscription process. \\
\textbf{Observation} & {
\tiny{
[1] RootWebArea \textquotesingle Grammarly plans - Premium, Business \& Student Writing Tools\textquotesingle  ~focused: True
\newline\hspace{10mm}[125] Generic \textquotesingle Grammarly plans - Premium, Business \& Student Writing Tools\textquotesingle 
\newline\tabto{0.5cm}[127] link \textquotesingle Privacy Policy\textquotesingle 
\newline\tabto{0.5cm}[130] link \textquotesingle Terms of Use\textquotesingle 
\newline\tabto{0.5cm}[133] link \textquotesingle Cookies\textquotesingle 
\newline\tabto{0.5cm}[136] link \textquotesingle TrustArc Privacy Notice\textquotesingle 
\newline\tabto{0.5cm}[587] link \textquotesingle Language\textquotesingle 
\newline\tabto{0.5cm}[1054] link \textquotesingle United States English\textquotesingle 
\newline\tabto{0.5cm}[592] link \textquotesingle Logout\textquotesingle 
\newline\tabto{0.5cm}[595] link \textquotesingle Invite\textquotesingle 
\newline\tabto{0.5cm}[598] link \textquotesingle Get the Extension\textquotesingle 
\newline\tabto{0.5cm}[128] link \textquotesingle Login\textquotesingle 
\newline\tabto{0.5cm}[601] link \textquotesingle Sign Up\textquotesingle 
\newline\tabto{0.5cm}[580] Generic \textquotesingle Plans\textquotesingle 
\newline\tabto{1cm}[583] div \textquotesingle Free\textquotesingle 
\newline\tabto{1.5cm}[586] link \textquotesingle Get Started\textquotesingle 
\newline\tabto{1cm}[590] Generic \textquotesingle Premium\textquotesingle 
\newline\tabto{1.5cm}[1098] Generic \textquotesingle Premium\textquotesingle  ~haspopup: submenu expanded: False
\newline\tabto{1.5cm}\hspace{10mm}[640] link \textquotesingle View Features\textquotesingle 
\newline\tabto{1.5cm}\hspace{10mm}[643] StaticText \textquotesingle Start Now!\textquotesingle 
\newline\tabto{1.5cm}\hspace{10mm}[1105] button \textquotesingle Choose Plan\textquotesingle 
\newline\tabto{1cm}[1089] Generic \textquotesingle Business\textquotesingle 
\newline\tabto{1.5cm}[649] link \textquotesingle View Pricing\textquotesingle 
\newline\tabto{1.5cm}[652] StaticText \textquotesingle Get Quote\textquotesingle 
\newline\tabto{1cm}[1092] Generic \textquotesingle Student\textquotesingle 
\newline\tabto{1.5cm}[646] link \textquotesingle View Pricing\textquotesingle 
\newline\tabto{1.5cm}[649] StaticText \textquotesingle Start Now!\textquotesingle 
\newline\tabto{0.5cm}[566] Generic \textquotesingle Write with confidence, wherever you work\textquotesingle 
\newline\tabto{1cm}[570] StaticText \textquotesingle Powerful writing tools for work, school, and everything in between.\textquotesingle 
\newline\tabto{1cm}[574] button \textquotesingle Learn More \&amp; Try it Free\textquotesingle 
\newline\tabto{0.5cm}[1052] table \textquotesingle 
\newline\tabto{1cm}[569] row \textquotesingle 
\newline\tabto{1.5cm}[574] rowheader \textquotesingle Free\textquotesingle 
\newline\tabto{1.5cm}[575] gridcell \textquotesingle 
\newline\tabto{1cm}[568] row \textquotesingle 
\newline\tabto{1.5cm}[573] rowheader \textquotesingle Premium\textquotesingle 
\newline\tabto{1.5cm}[574] gridcell \textquotesingle 
\newline\tabto{1cm}[567] row \textquotesingle 
\newline\tabto{1.5cm}[572] rowheader \textquotesingle Business\textquotesingle 
\newline\tabto{1.5cm}[573] gridcell \textquotesingle 
\newline\tabto{1cm}[566] row \textquotesingle 
\newline\tabto{1.5cm}[571] rowheader \textquotesingle Student\textquotesingle 
\newline\tabto{1.5cm}[572] gridcell \textquotesingle 
}}
\\
\bottomrule[\heavyrulewidth]
\end{tabularx}
}

    \caption{A selected full sample of a generated novel, out-of-domain synthetic training example in $\mathcal{D}_{\textsc{Out-of-Domain}}$.}
    \label{table:fulloutofdomainexample}
    \end{table}
}
\newcommand{\experimentstable}{
    \begin{table}[t!]
    \small
    \centering
    \begin{tabular}{l>{\raggedleft\arraybackslash}p{.13\textwidth}rr}
    \toprule[\heavyrulewidth]
    \textbf{Agent Model} & \textbf{Functional Correctness} & \textbf{Capability} & $\textbf{VERTEX}_{\textbf{DTW}}$ \\ \midrule
    \multicolumn{4}{l}{\textit{Baseline Agents}} \\ \hline
    Trivial Agent & 4.68 & 0.00 & - \\
    Base Agent Model ($\mathcal{M}$) & 7.14 & 15.44 & 0.35 \\ \hline
    {\textit{Self-Improved Agents}} \\ \hline
    Agent Model Fine-Tuned on Mixture A ($\mathcal{M}_A$) & 8.87 & \textbf{19.12} & \textbf{0.38}  \\
    Agent Model Fine-Tuned on Mixture B ($\mathcal{M}_B$) & \textbf{9.36} & \textbf{19.12} & 0.35  \\
    Agent Model Fine-Tuned on Mixture C ($\mathcal{M}_C$) & 6.16 & 16.91 & 0.28  \\ \hline
    {\textit{Iterative Self-Improved Agents}} \\ \hline
    Agent Model 2x Fine-Tuned on
    Mixture A ($\mathcal{M}_A^2$) & 8.37 & 16.91 & 0.37  \\

    \bottomrule[\heavyrulewidth]
    \end{tabular}
    \caption{Evaluation metrics on WebArena for baseline agents and self-improved agent models.}
    \label{table:experiments}
    \end{table}
}
\newcommand{\iterativefilteringtable}{
    \begin{table}[h]
    \small
    \centering
    \begin{tabular}{lrrrrr}
    \toprule[\heavyrulewidth]
    \textbf{Set of Trajectories} & \textbf{\#} & \textbf{Accuracy} & \textbf{F1} & \textbf{Precision}  & \textbf{Recall} \\ \midrule
    All Trajectories & 812 & 0.071 & 0.133 & 0.071 & 1.000 \\
    Plausible Trajectories $\mathcal{P}^1$ & 58 & \textbf{0.919} & \textbf{0.431} & \textbf{0.431} & \textbf{0.431}  \\
    Plausible Trajectories $\mathcal{P}^2$ & \textbf{131} & 0.825 & 0.317 & 0.252 & 0.429  \\

    \bottomrule[\heavyrulewidth]
    \end{tabular}
    \caption{Metrics on the proportion of trajectories that successfully completed the task in the set of plausible trajectories kept in $\mathcal{P}$ after filtering out low-quality trajectories for each iterative round of self-improvement. On the second round of self-improvement, we keep 131 plausible trajectories making our potential synthetic training dataset larger, however, the accuracy and P/R/F1 metrics indicate it would be a lower quality dataset to fine-tune on.}
    \label{table:iterativefiltering}
    \end{table}
}
\newcommand{\capabilityanalysistable}{
{\small

\begin{longtable}[H!]{m{.13\textwidth}m{.12\textwidth}m{.30\textwidth}m{.30\textwidth}}

\toprule[\heavyrulewidth]
\textbf{Agent Model} & \textbf{Net Change} & \textbf{Capabilities Acquired} & \textbf{Capabilities Lost} \\
\hline
$\mathcal{M}_A$ & +5 &
{\flushleft
\scriptsize
\begin{enumerate}[noitemsep,leftmargin=*]
    \item Tell me the count of comments that have received more downvotes than upvotes for the user who made the latest post on the \texttt{\tiny \{\{forum\}\}} forum.
    \item Find a subreddit focused on topics related to \texttt{\tiny \{\{topic\}\}}, and post my question, "\texttt{\tiny \{\{question\}\}}" there
    \item Measure distance between \texttt{\tiny \{\{location/address\_1\}\}} and \texttt{\tiny \{\{location/address\_2\}\}} by walking
    \item Tell me the coordinates of \texttt{\tiny \{\{location\}\}} in DD format
    \item Show me the "\texttt{\tiny \{\{product\}\}}" listings by \texttt{\tiny \{\{sorting\_order\}\}}.
    \item Open my latest updated issue that has keyword "\texttt{\tiny \{\{keyword\}\}}" in its title to check if it is closed
    \item Reply to \texttt{\tiny \{\{position\_description\}\}} with my comment "\texttt{\tiny \{\{content\_description\}\}}"
    \item Tell me the distance to drive from Carnegie Mellon University to the top computer science school in massachusetts
    \item How many commits did \texttt{\tiny \{\{user\}\}} make on \texttt{\tiny \{\{date\}\}} in total?
\end{enumerate}
}
& 
{\flushleft
\scriptsize
\begin{enumerate}[noitemsep,leftmargin=*]
    \item Tell me the total cost of my latest \texttt{\tiny \{\{status\}\}} order?
    \item Checkout merge requests assigned to me
    \item Today is 6/12/2023. Tell me how many fulfilled orders I have \texttt{\tiny \{\{period\}\}}, and the total amount of money I spent.
    \item Subscribe to the newsletter of OneStopMarket
\end{enumerate}
}
\\
$\mathcal{M}_B$ & +5 &
{\flushleft
\scriptsize
\begin{enumerate}[noitemsep,leftmargin=*]
    \item Tell me the count of comments that have received more downvotes than upvotes for the user who made the latest post on the \texttt{\tiny \{\{forum\}\}} forum.
    \item What is the minimum travel time by car from \texttt{\tiny \{\{location1\}\}} to \texttt{\tiny \{\{location2\}\}}?
    \item Find a subreddit focused on topics related to \texttt{\tiny \{\{topic\}\}}, and post my question, "\texttt{\tiny \{\{question\}\}}" there
    \item See all public projects
    \item Set my gitlab status as \texttt{\tiny \{\{status\}\}}.
    \item Show me the route and driving time from \texttt{\tiny \{\{city1\}\}} to \texttt{\tiny \{\{city2\}\}}
    \item Ask for advice about \texttt{\tiny \{\{issue\}\}} in a subreddit for relations
    \item Show me the "\texttt{\tiny \{\{product\}\}}" listings by \texttt{\tiny \{\{sorting\_order\}\}}.
    \item Reply to \texttt{\tiny \{\{position\_description\}\}} with my comment "\texttt{\tiny \{\{content\_description\}\}}"
    \item Show me the way from \texttt{\tiny \{\{location\}\}} to the home stadium of \texttt{\tiny \{\{sport\_team\}\}} \texttt{\tiny \{\{time\}\}}
\end{enumerate}
}
& 
{\flushleft
\scriptsize
\begin{enumerate}[noitemsep,leftmargin=*]
    \item Checkout merge requests assigned to me
    \item Today is 6/12/2023. Tell me how many fulfilled orders I have \texttt{\tiny \{\{period\}\}}, and the total amount of money I spent.
    \item Subscribe to the newsletter of OneStopMarket
    \item Show me the command to clone \texttt{\tiny \{\{repo\}\}} with SSH.
    \item Show me the \texttt{\tiny \{\{info\}\}} for order number \texttt{\tiny \{\{order\_number\}\}}.
\end{enumerate}
} \\
\multicolumn{4}{c}{\textit{Continued on next page$\ldots$}} \\
$\mathcal{M}_C$ & +2 &
{\flushleft
\scriptsize
\begin{enumerate}[noitemsep,leftmargin=*]
    \item Fork \texttt{\tiny \{\{repo\}\}}.
    \item Tell me the count of comments that have received more downvotes than upvotes for the user who made the latest post on the \texttt{\tiny \{\{forum\}\}} forum.
    \item Which US states border \texttt{\tiny \{\{state\}\}}?
    \item What is the minimum travel time by car from \texttt{\tiny \{\{location1\}\}} to \texttt{\tiny \{\{location2\}\}}?
    \item See all public projects
    \item I previously ordered some \texttt{\tiny \{\{product\}\}} \texttt{\tiny \{\{time\}\}} and later cancelled. Can you reorder it for me?
    \item Today is 3/15/2023, generate a \texttt{\tiny \{\{report\}\}} \texttt{\tiny \{\{time\_span\}\}}
    \item Ask for advice about \texttt{\tiny \{\{issue\}\}} in a subreddit for relations
    \item Show me the "\texttt{\tiny \{\{product\}\}}" listings by \texttt{\tiny \{\{sorting\_order\}\}}.
    \item Pull up the description page of \texttt{\tiny \{\{location\}\}} on Map
    \item Reply to \texttt{\tiny \{\{position\_description\}\}} with my comment "\texttt{\tiny \{\{content\_description\}\}}"
    \item Edit my post on \texttt{\tiny \{\{post\}\}} by adding a line to the body that says "\texttt{\tiny \{\{content\}\}}"
    \item Tell me who has made the most contributions, in terms of number of commits, to the \texttt{\tiny \{\{repo\}\}} project
    \item List the top \texttt{\tiny \{\{n\}\}} search terms in my store
\end{enumerate}
}
& 
{\flushleft
\scriptsize
\begin{enumerate}[noitemsep,leftmargin=*]
    \item How many commits did \texttt{\tiny \{\{user\}\}} make to \texttt{\tiny \{\{repo\}\}} on \texttt{\tiny \{\{date\}\}}?
    \item Checkout merge requests assigned to me
    \item What is the estimated driving time between \texttt{\tiny \{\{city1\}\}} and \texttt{\tiny \{\{city2\}\}}?
    \item Open the thread of a trending post on the forum "\texttt{\tiny \{\{subreddit\}\}}" and subscribe.
    \item Today is 6/12/2023. Tell me how many fulfilled orders I have \texttt{\tiny \{\{period\}\}}, and the total amount of money I spent.
    \item Find the \texttt{\tiny \{\{space\}\}} around \texttt{\tiny \{\{location\}\}}
    \item What are the main criticisms of this product? Please extract the relevant sentences.
    \item Subscribe to the newsletter of OneStopMarket
    \item Show me the command to clone \texttt{\tiny \{\{repo\}\}} with SSH.
    \item I want to browse the products in the \texttt{\tiny \{\{category\}\}} category
    \item Show me the \texttt{\tiny \{\{info\}\}} for order number \texttt{\tiny \{\{order\_number\}\}}.
    \item Tell me the total cost of my latest \texttt{\tiny \{\{status\}\}} order?
\end{enumerate}
} \\
\bottomrule[\heavyrulewidth]

    \caption{Capabilities acquired and lost compared to the base agent model $\mathcal{M}$, along with the net change in the total number of capabilities demonstrated, for each self-improved fine-tuned agent model.}
    \label{table:capabilityanalysis}
\end{longtable}
}
\
}
\newcommand{\fullvertextable}{
    \begin{table}[H]
    \scriptsize
    \centering
    \begin{tabular}{lrrr}
    \toprule[\heavyrulewidth]
    \textbf{Agent Model} & $\textbf{VERTEX}_{\textbf{DTW}}$ & $\textbf{VERTEX}_{\textbf{DTW-bycap}}$ & $\textbf{VERTEX}_{\textbf{DTW-notrivial}}$ \\ \midrule
    \multicolumn{4}{l}{\textit{Baseline Agents}} \\ \hline
    Base Agent Model  ($\mathcal{M}$) & 0.35 & 0.40 & 0.38 \\ \hline
    {\textit{Self-Improved Agents}} \\ \hline
    Agent Model Fine-Tuned on Mixture A ($\mathcal{M}_A$) & \textbf{0.38} & \textbf{0.42} & 0.42  \\
    Agent Model Fine-Tuned on Mixture B ($\mathcal{M}_B$) & 0.35 & 0.40 & 0.40  \\
    Agent Model Fine-Tuned on Mixture C ($\mathcal{M}_C$) & 0.28 & 0.33 & 0.34  \\ \hline
    {\textit{Iterative Self-Improved Agents}} \\ \hline
    Agent Model 2x Fine-Tuned on
    Mixture A ($\mathcal{M}_A^2$) & 0.37 & 0.41 & \textbf{0.43}  \\

    \bottomrule[\heavyrulewidth]
    \end{tabular}
    \caption{Variants of the $\text{VERTEX}_{\text{DTW}}$ score metric: 1) computed over all trajectories 2) weighting the trajectories by capability 3) weighting the trajectories by capability and filtering out trajectories for trivial tasks.}
    \label{table:fullvertex}
    \end{table}
}

\title{Large Language Models Can Self-Improve \newline At Web Agent Tasks}

\author{%
  ~~~~~~~~~~~~~~~~Ajay Patel \textsuperscript{\textdagger}~~~~~~~~~Markus Hofmarcher\textsuperscript{\textdaggerdbl~\textsection}~~~~~~~~~Claudiu Leoveanu-Condrei\textsuperscript{{\textdaggerdbl}}\\ ~~~~~~~~~~~~~\textbf{Marius-Constantin Dinu\textsuperscript{\textdaggerdbl~\textsection}~~~~~~~~~Chris Callison-Burch\textsuperscript{\textdagger}~~~~~~~~~Sepp Hochreiter\textsuperscript{\textbardbl~\textsection}} \\
  University of Pennsylvania\textsuperscript{\textdagger}~~~~~~~~~ExtensityAI\textsuperscript{\textdaggerdbl}~~~~~~~~~Johannes Kepler University Linz\textsuperscript{\textsection}~~~~~~~~~NXAI\textsuperscript{\textbardbl} \\
  ~~~~~~~~~~~~~~~~~~~~~\texttt{\{ajayp,ccb\}@upenn.edu},~~~~ \texttt{\{markus,leo\}@extensity.ai} \\
  ~~~~~~~~~~~~~~~~~~~~~~~~~~~~~~~~~~~~~~~~~~~~~~~~\texttt{\{dinu,hochreit\}@ml.jku.at}
}

\iclrfinalcopy

\begin{document}

\maketitle

\begin{abstract}
Training models to act as agents that can effectively navigate and perform actions in a complex environment, such as a web browser, has typically been challenging due to lack of training data. Large language models (LLMs) have recently demonstrated some capability to navigate novel environments as agents in a zero-shot or few-shot fashion, purely guided by natural language instructions as prompts. Recent research has also demonstrated LLMs have the capability to exceed their base performance through self-improvement, i.e. fine-tuning on data generated by the model itself. In this work, we explore the extent to which LLMs can self-improve their performance as agents in long-horizon tasks in a complex environment using the WebArena benchmark. In WebArena, an agent must autonomously navigate and perform actions on web pages to achieve a specified objective. We explore fine-tuning on three distinct synthetic training data mixtures and achieve a 31\% improvement in task completion rate over the base model on the WebArena benchmark through a self-improvement procedure. We additionally contribute novel evaluation metrics for assessing the performance, robustness, capabilities, and quality of trajectories of our fine-tuned agent models to a greater degree than simple, aggregate-level benchmark scores currently used to measure self-improvement.
\end{abstract}


\section{Introduction}
\label{sec:introduction}

Large language models (LLMs) have demonstrated impressive capabilities in a variety of natural language processing (NLP) tasks such as summarization and question answering \citep{gpt2,t5,gpt3} through zero-shot and few-shot prompting techniques \citep{instructgpt,flan}. However, prompting techniques alone are insufficient to enable LLMs to act as agents and navigate environments in order to solve complex, multi-step, long-horizon tasks \citep{react}. Fine-tuning LLMs to perform such tasks is also infeasible due to the scarcity of training data suitable for these tasks. Acquiring data for sequential decision-making and complex interactions is not only time-consuming, but also costly. Additionally, automatic evaluation of trajectories (or sequences of actions) taken by an agent is also difficult \citep{symbolicai}.
The absence of metrics that accurately capture the efficacy of each step in a sequence complicates the assessment of incremental improvements or degradations in an agent's performance. 

A number of proposed self-improvement techniques have demonstrated that LLMs can use zero-shot and few-shot prompting to achieve performance above the baseline without any additional supervised training data \citep{llmscanselfimprove,selfplay}. In place of supervised data as a learning signal, many of these techniques use a self-critique technique \citep{selfverification,selfrewarding}, or obtain a critique through interactions with tools or environments \citep{environmentcritique}. While self-improvement techniques have shown promise on standard NLP benchmark tasks like machine translation or question answering \citep{unsupervisedmt,llmscanselfimprove,selfplay}, their efficacy has not yet been thoroughly investigated for long-horizon tasks that require multi-step interactions with a complex and realistic environment.

WebArena \citep{webarena} is a recently proposed benchmark wherein an LLM agent is required to solve tasks using a web browser. One example WebArena task is to use the OpenStreetMap website to answer the question ``\textit{What is the minimum travel time by car from CMU to University of Pittsburgh?}''. Such a task requires an agent to complete a sequence of steps on the website, including entering a start location, entering a destination location, submitting a form, and then, reasoning over the result. The sequence of steps selected by an agent is called a {\it trajectory}.  Unlike existing benchmarks, WebArena tasks are realistic and diverse, require dynamic interaction,  and require navigating a complex environment. The baselines presented by \cite{webarena} demonstrate that while LLMs are capable of interacting with this environment, even the strongest baseline, GPT-4 \citep{gpt4}, is only able to solve \textasciitilde 14\% of the tasks. This demonstrates that WebArena is a challenging benchmark even for the strongest frontier models \citep{lmysys}.

In this paper, we introduce new techniques that allow LLM agents to better perform complex, multi-step tasks via self-improvement. We detail different strategies for self-improvement that all involve fine-tuning the LLM agent on its own generations (synthetic data) and inducing a signal for learning by employing unsupervised techniques like self-critique to selectively filter training examples. To better understand the effect of our self-improvement, we introduce two auxiliary metrics: 1) a measure to analyze capabilities acquired and lost by the agent and 2) an extension of the VERTEX score \citep{symbolicai} to measure the quality of variable-length agent trajectories. These metrics allow finer-level assessment of improvements and degradations than aggregate-level benchmark scores.

\mainfig

In summary, our contributions are:
\begin{itemize}
    \item We propose and detail procedures for collecting and generating synthetic training examples for complex, multi-step tasks involving interaction with an environment.  We explore collecting in-domain synthetic examples of trajectories as well as generating synthetic examples of solution trajectories for novel, out-of-domain tasks.
    \item We show that the performance of LLM agents  improves after fine-tuning on this synthetic data, demonstrating that self-improving techniques work for a new class of tasks.  We analyze three synthetic training data mixtures and find all three mixtures improve performance, with the best performing mixture yielding a 31\% improvement over the base LLM agent on the WebArena benchmark.
    \item We propose auxiliary metrics to understand the effect self-improvement has with respect to acquiring new capabilities and to evaluate variable-length trajectories produced by agents through an extension of the VERTEX score. These metrics provide nuanced insights not captured by aggregate-level benchmark scores currently used to evaluate self-improvement, allowing us to better assess the effect self-improvement has on multiple dimensions: performance, robustness, capability acquisition, and the quality of generated trajectories.
\end{itemize}

\section{Synthetic Data Collection and Generation}
\label{sec:data}

Self-improvement techniques for large language models typically involve using the model's own generations to create synthetic few-shot examples \citep{unsupervisedmt} or synthetic fine-tuning data \citep{llmscanselfimprove}. These techniques amplify knowledge, correct behaviors, and introduce regularization \citep{selfdistill2}, often leading to an overall boost in performance. The self-generated examples are often filtered, post-edited, or ranked with a set of unsupervised techniques such as self-critique to introduce a signal for learning and improvement \citep{selfverification,bootstrapmtexamples,selfplay,selfrewarding}. For multi-step agent tasks, the environment itself can additionally provide the LLM agent a way to detect failure in a fully unsupervised manner, which provides another useful signal for learning \citep{environmentcritique,rft,trialanderror}.

Using the WebArena benchmark \citep{webarena}, we define and experiment with both in-domain synthetic training examples and out-of-domain synthetic training examples for web agent tasks, and fine-tune on three different synthetic data mixtures: \textbf{Mixture A} (in-domain synthetic examples only), \textbf{Mixture B}  (both in-domain and out-of-domain synthetic examples), and \textbf{Mixture C} (out-of-domain synthetic examples only). Figure \ref{fig:main} illustrates our process.

\paragraph{\textbf{In-Domain Synthetic Data}:} For all tasks in WebArena, we collect an initial set of trajectories using the base model. We filter out any trajectories where the model self-detected failure (self-critique) or failure was detectable in the environment and keep the remainder. We denote the remaining set of trajectories as \textit{plausible trajectories}, where the model may or may not have completed the task successfully. Since lower-quality trajectories where the model outright failed to complete the task have been filtered out through self-detection, we hypothesize this remaining higher-quality set of plausible trajectories can serve as reasonably high-quality \textit{in-domain synthetic examples} for fine-tuning. Similar to the self-improvement prior work we discuss earlier, the collection of this data is completely unsupervised and no ground-truth labels are utilized for filtering and selection.

\paragraph{\textbf{Out-of-Domain Synthetic Data}:} We also evaluate whether the base model can generate completely novel tasks, objectives, web pages, and solution trajectories that can serve as useful training examples. We use the plausible trajectories as few-shot examples in a prompt for the base model to generate completely new tasks along with potential solution trajectories. To ensure the model generates examples with sufficient diversity and to improve generalization, we prompt the model to generate \textit{out-of-domain synthetic examples} that are dissimilar from existing tasks and objectives as well as generate tasks for different websites than the set of 6 websites covered by the WebArena benchmark.

\subsection{In-Domain Synthetic Data Collection}

The WebArena environment can be formulated as a partially observable Markov decision process: $\mathcal{E} = \langle \mathcal{S}, \mathcal{A}, \mathcal{O}, \mathcal{T} \rangle$, where $\mathcal{S}$ represents the state space, $\mathcal{A}$ represents the action space, $\mathcal{O}$ represents the observation space, and $\mathcal{T}: \mathcal{S} \times \mathcal{A} \rightarrow \mathcal{S}$ is the deterministic transition function \citep{webarena}. An agent model $\mathcal{M}$ produces a next action $a_t \in \mathcal{A}$ provided an objective represented by some natural language intent $\mathbf{i}$, the current observation $o_t \in \mathcal{O}$, and the previous action taken $a_{t-1} \in \mathcal{A}$: $(\mathbf{i}, o_t, a_{t-1})$. This continues for $T$ time steps until the agent produces a stop action or the environment produces an error or stop condition. The model $\mathcal{M}$ we select for our experiments is the Qwen-1.5-72B-Chat model \citep{qwen}, which at the time of this work is a highly ranked\footnote{\scriptsize \url{https://chat.lmsys.org/?leaderboard}}  and competitive open source LLM \citep{lmysys} that is accessible for fine-tuning. Further choice of inference parameters and other configuration details can be found in Appendix \ref{sec:appendix_training_and_inference_details}.

Given this definition, we propose a procedure for sampling a set of in-domain synthetic training examples $\mathcal{D}_{\textsc{In-Domain}}$ where each training example is structured as $(\mathbf{i}, o_t, a_{t-1}) \rightarrow a_t$. These examples are sampled from a filtered set of trajectories collected by an initial run of the base agent model $\mathcal{M}$ over all tasks in WebArena:

\begin{algorithm}[H]
\caption{~Collect In-Domain Synthetic Training Examples $\mathcal{D}_{\textsc{In-Domain}}$}
\label{alg:in-domain-examples}
\begin{algorithmic}[1]
\small
\Require WebArena environment $\mathcal{E}$ and base agent model $\mathcal{M}$
\Ensure A set of in-domain synthetic training examples $\mathcal{D}_{\textsc{In-Domain}}$
\State Initialize $\mathcal{P} \leftarrow \emptyset$ \CommentTriangle{Set of plausible trajectories}
\For{$\mathbf{i}$ in WebArena benchmark}
    \State Initialize trajectory $\mathcal{X} \leftarrow \emptyset$
    \State Initialize observation $o_{0} \leftarrow \Call{\small InitialObservation}{\mathcal{E}, \mathbf{i}}$
    \State Initialize action $a_{-1} \leftarrow \text{null}$
    \For{$t = 0$ to $T$}
        \State $a_t \leftarrow \Call{\small RunAgent}{\mathcal{M}, \mathbf{i}, o_t, a_{t-1}}$
        \State Append $(\mathbf{i}, o_t, a_{t-1}, a_t)$ to $\mathcal{X}$
        \If{$a_t = \texttt{stop}~\textbf{or}~\Call{\small EnvironmentError}{\mathcal{E},a_t, o_{t+1}}$}
            \State \textbf{break}
        \EndIf
        \State $o_{t+1} \leftarrow \mathcal{T}(o_t, a_t)$ \CommentTriangle{Observe updated state}
    \EndFor
    \If{\textbf{not} \Call{\small SelfCritique}{$\mathcal{X}$} \textbf{and} \textbf{not} \Call{\small IsRefusal}{$\mathcal{X}$} \textbf{and} \textbf{not} \Call{\small HasError}{$\mathcal{X}$}}
        \State Append $\mathcal{X}$ to $\mathcal{P}$ \CommentTriangle{Filter out low-quality trajectories \newline \text{\textcolor{white}{\tiny .}} \hfill to only keep plausible trajectories}
    \EndIf
\EndFor
\State Initialize $\mathcal{D}_{i}, \mathcal{D}_{f}, \mathcal{D}_{int} \leftarrow \emptyset$ \CommentTriangle{Set of initial steps, final steps, intermediate steps}
\For{$\mathcal{X}$ in $\mathcal{P}$}
        \State Append $\mathcal{X}_0$ to $\mathcal{D}_{i}$
        \State Append $\mathcal{X}_{T}$ to $\mathcal{D}_{f}$
        \For{$t = 1$ to $T-1$}
            \State Append $\mathcal{X}_{t}$ to $\mathcal{D}_{int}$
        \EndFor        
\EndFor
\State $\mathcal{D}_{\textsc{In-Domain}} \leftarrow  \Call{\small RandSample}{\mathcal{D}_i, |\mathcal{D}_i|} \cup \Call{\small RandSample}{\mathcal{D}_f, |\mathcal{D}_i|} \cup \Call{\small RandSample}{\mathcal{D}_{int}, 2 * |\mathcal{D}_i|}$
\State \Return $\mathcal{D}_{\textsc{In-Domain}}$
\end{algorithmic}
\end{algorithm}

We filter out low-quality trajectories where the model produced a generation stating the task to be ``impossible'' or that it ``cannot'' make progress (a form of self-critique). Additionally, we filter out any trajectories where the model produced \texttt{\small stop[N/A]}, \texttt{\small stop[No ...]}, or \texttt{\small stop[]}, indicating when the model may have refused to provide an answer. Finally, we also filter out any trajectories where the WebArena environment encountered an error or the model failed to produce a valid, parsable generation. The final dataset of synthetic examples is balanced by randomly sampling an equal number of initial steps ($t=0$), final steps ($t=T$), and intermediate steps ($t=1\ldots(T-1)$) from the plausible trajectories in $\mathcal{P}$. In Table \ref{table:indomainfiltering}, we display how effective this unsupervised filtering process is by measuring the accuracy, precision, and recall of the 58 remaining trajectories kept in $\mathcal{P}$ from the 812 total trajectories to assess the proportion of correct/incorrect examples in $\mathcal{D}_{\textsc{In-Domain}}$.

\indomainfilteringtable

\subsection{Out-of-Domain Synthetic Data Generation}

Using examples from $\mathcal{D}_{\textsc{In-Domain}}$ as seed examples, we prompt our base LLM $\mathcal{M}$ to synthetically generate completely novel tasks, objectives, web pages, and solution trajectories to produce $\mathcal{D}_{\textsc{Out-of-Domain}}$.

\begin{algorithm}[H]
\caption{~Generate Out-of-Domain Synthetic Training Examples $\mathcal{D}_{\textsc{Out-of-Domain}}$}
\label{alg:out-of-domain-examples}
\begin{algorithmic}[1]
\small
\Require Base LLM model $\mathcal{M}$ and $\mathcal{D}_{\textsc{In-Domain}}$
\Ensure A set of out-of-domain synthetic training examples $\mathcal{D}_{\textsc{Out-of-Domain}}$
\State Initialize $\mathcal{D}_{\textsc{Out-of-Domain}} \leftarrow \emptyset$ \CommentTriangle{Set of out-of-domain synthetic training examples}
\State Initialize $\mathcal{I} \leftarrow \{\mathbf{i} \mid \mathbf{i} \in \text{WebArena benchmark}\}$ 
\CommentTriangle{Set of 812 objectives in WebArena}
\State Initialize $\mathcal{I}^* \leftarrow \emptyset$ 
\CommentTriangle{Set of previously generated objectives}
\For{$j = 1$ to $|\mathcal{D}_{\textsc{In-Domain}}|$}
    \While {true}
        \State $\mathbf{i}^* \leftarrow \Call{\small GenerateObjective}{\mathcal{M}, \Call{\scriptsize RandSample}{\mathcal{I}, 2} \cup \Call{\scriptsize RandSample}{\mathcal{I}^*, 2}}$
        \If{$\text{max}(\text{sim}(\mathbf{i}^*, \mathcal{I}^*)$) < 0.70} \CommentTriangle{Ensure generated objectives are diverse}
            \State Append $\mathbf{i}^*$ to $\mathcal{I}^*$
            \State \textbf{break}
        \EndIf
    \EndWhile
    \State $\mathbf{p}^* \leftarrow \Call{\small GeneratePlan}{\mathcal{M}, \mathbf{i}^*}$\CommentTriangle{Generate an outline of a hypothetical solution trajectory}
    \State $k \leftarrow \Call{\small RandChoice}{\{1, \ldots, |\mathbf{p}^*| \}}$\CommentTriangle{Randomly select one of the steps in the plan, weighted\newline \text{\textcolor{white}{\tiny .}} \hfill to equally balance initial, final, and intermediate steps}
    \State $a_{t-1}^*,a_{t}^* \leftarrow \Call{\small GenerateActions}{\mathcal{M}, \Call{\scriptsize RandSample}{\mathcal{D}_{\textsc{In-Domain}}, 2}, \mathbf{i}^*, \mathbf{p}^*, k}$
    \State $o_{t}^* \leftarrow \Call{\small GenerateObservation}{\mathcal{M}, \Call{\scriptsize RandSample}{\mathcal{D}_{\textsc{In-Domain}}, 2}, \mathbf{i}^*, \mathbf{p}^*, k}$
    \State Append $(\mathbf{i}^*, o_t^*, a_{t-1}^*, a_t^*)$ to $\mathcal{D}_{\textsc{Out-of-Domain}}$
\EndFor
\State \Return $\mathcal{D}_{\textsc{Out-of-Domain}}$
\end{algorithmic}
\end{algorithm}

When generating new objectives, we use 4 few-shot examples (two objectives sampled from tasks in WebArena and two sampled from previously generated objectives). We use 2 few-shot examples when generating previous actions, next actions, and observations (web pages in the form of accessibility trees). We use a temperature of 1.0 and set top-p to 1.0  during generation. Detailed information on the prompts used for generating $\mathcal{D}_{\textsc{Out-of-Domain}}$ can be found in Appendix \ref{sec:appendix_prompts}. When generating novel objectives, we specifically prompt the model to generate objectives that are dissimilar to the example objectives to encourage out-of-domain generations. We also ensure each novel objective has $<0.70$ cosine similarity with any objective previously generated using the \texttt{\small all-distilroberta-v1} sentence similarity model \citep{sentencetransformers,roberta,distilroberta} to promote diversity.  Table \ref{table:outofdomainexamples} gives examples of out-of-domain objectives that our method generated.

\outofdomainexamplestable
\section{Evaluation}
\label{sec:evaluation}

We perform evaluation using the standard metrics proposed by the WebArena benchmark like functional correctness \citep{webarena} as well as evaluate with new auxiliary metrics we propose that give more nuanced insight into an agent's performance.

\subsection{Functional Correctness Score} Functional correctness is the standard metric proposed by the WebArena benchmark that is a simple binary task completion score (0 or 1) averaged over all 812 tasks in the benchmark.

\subsection{Capability Score (New)}
\label{sec:capabilityscore}

While WebArena contains 812 unique task instances, these 812 tasks are instantiated using natural language intent templates like ``What is the minimum travel time by car from \texttt{\small \{\{location1\}\}} to \texttt{\small \{\{location2\}\}}?''. Therefore, many tasks actually test the same \textit{capability}. Aggregate-level metrics like the functional correctness score may be misleading since improvements may only be due to the model becoming more robust at solving capabilities it already could solve versus demonstrating the ability to solve new capabilities that were previously unsolvable. There are 241 unique templates in WebArena that are used to instantiate 812 tasks. Moreover, some of these templates are simple paraphrases of each other. For example, ``What is the estimated driving time between \texttt{\small \{\{city1\}\}} and \texttt{\small \{\{city2\}\}}?'' is a paraphrase of the prior template. Using a sentence similarity model \footnote{We use the \texttt{\scriptsize all-distilroberta-v1} sentence similarity model \citep{distilroberta}.}, we iteratively group these templates into a set of unique capabilities. Each template is grouped with any existing capability if it has a similarity of >~0.60 with any template in the group, otherwise the template is added to a new capability group. This results in 136 unique capabilities (see Appendix \ref{sec:appendix_capabilities}). A model receives a score of 1 for each capability group with at least one successful task completed, otherwise it receives a score of 0 \footnote{Since we do not count trivial tasks as a successful completion, a single successful completion of a capability provides sufficient evidence of acquisition. We discuss robustness and consistency separately in Section \ref{sec:discussion}.}. The capability score is then the averaged over all 136 capabilities.

We note, however, that a number of tasks in the WebArena benchmark are trivial tasks and can be solved by a trivial baseline agent or weak model that performs no actions and only immediately exits by always generating \texttt{\small stop [N/A]}. In the capability score computation, we do not count such trivial tasks as evidence a model can perform the capability as these are degenerate cases of the capability.

\subsection{$\text{VERTEX}_{\text{DTW}}$ Score (New)}

Both functional correctness and the capability score only evaluate task completion, however, they do not assess the quality of entire trajectories, therefore, a measure that is sensitive to incremental improvements and degradations in trajectories, independent of task completion, is desirable.\textbf{} We extend the recently proposed VERTEX score \citep{symbolicai}, which measures the similarity of two relational trajectories by using embeddings to compare node distributions within a computational graph. The VERTEX score integrates the semantic meaning across the distributional path by computing at each node the cross-similarity between the generated embeddings and embeddings sampled from a reference distribution. An ideal reference distribution would be ground-truth reference trajectories produced by humans for all of the WebArena tasks. In absence of this, we use a larger, stronger model, GPT-4 \citep{gpt4}, to collect three reference trajectories for each task.

One obstacle to the straightforward application of the VERTEX score is the assumption that both trajectories are of the same length. Agents operating in complex environments, however, are not constrained to a fixed-length for the trajectories they produce. Therefore, we propose modification in the computation of the VERTEX score that enables comparison of sequences with different lengths. Our extension consists of an additional alignment step prior to calculating the VERTEX score for the aligned trajectories. 
First, we embed all steps of a trajectory $\mathcal{X}$ as $e_t=f(o_t, a_t) \in R^{d}$, where $f$ is an embedding model\footnote{We use the \texttt{\scriptsize all-mpnet-base-v2} embedding model \citep{Song:20mpnet}.} with embedding dimension $d$. 
The embedding model $f$ is independent of both the model that generated the reference trajectories as well as the model that generated the test trajectories. 
Then, we use \textit{Dynamic Time Warping} (DTW) \citep{Berndt:94dtw} to align two embedded trajectories $\tilde{\mathcal{X}}_m=(e_0, \dots, e_i, \dots, e_m) \in R^{m \times d}$ and $\tilde{\mathcal{X}}_n=(e_0, \dots, e_j, \dots, e_n) \in R^{n \times d}$ with length $m$ and $n$, respectively. 
Consequently, we refer to our proposed measure as $\text{VERTEX}_{\text{DTW}}$.
DTW returns an alignment path $\nu$ of length $T$, where each $e_i \in \tilde{\mathcal{X}}_m$ is aligned with a corresponding $e_j \in \tilde{\mathcal{X}}_n$, preserving the order in their respective trajectory.
This order preservation occurs because once a node is matched, it is excluded from potential new matches, maintaining the integrity of the temporal alignment.
As a scoring function for DTW, we choose cosine distance.
In addition to the alignment step, we introduce a linear distance decay factor that decreases the contribution of aligned embeddings if they are far apart in the original trajectories.
Once two trajectories are aligned, we compute the VERTEX score by Eq. (4) in \cite{symbolicai} with the addition of the distance decay.
Therefore, the $\text{VERTEX}_{\text{DTW}}$ score is computed as:

\begin{equation}
s(\tilde{\mathcal{X}}_{\text{ref}}, \tilde{\mathcal{X}}_{\text{test}}, \nu) := \frac{1}{T} \int_{t_0}^{t_T} \big [\min(\max(0,\frac{1}{1 + |i_{\nu_t}-j_{\nu_t}|} \ \widetilde{\text{MMD}^2}(e^{\nu_t}_{\text{ref}},e^{\nu_t}_{\text{test}})-z_{\mathrm{rand}}),1 ) \big ] dt,
\end{equation}

where $i_{\nu_t}$ and $j_{\nu_t}$ are the position indices in the alignment path $\nu$ at time $t$, $\mathcal{\Tilde{X}}_{\text{ref}}$ and $\mathcal{\Tilde{X}}_{\text{test}}$ are aligned trajectories of embeddings from the reference set and the model under test, respectively, and $z_{\mathrm{rand}}$ is a baseline correction from a random baseline\footnote{We use the trivial agent implementation described in Section \ref{sec:baselines} for baseline correction in our computation.}.
Furthermore, if we have multiple reference sequences for a given task, we compute the $\text{VERTEX}_{\text{DTW}}$ score for every reference sequence and choose the maximum score, under the assumption that they describe different paths for solving the task. 
\section{Experiments}
\label{sec:experiments}

We perform a number of experiments fine-tuning agent models on the synthetic training data mixtures we discuss in Section \ref{sec:data} and assess the extent to which the agent model has self-improved over base agent model $\mathcal{M}$ with our evaluation metrics. Table \ref{table:experiments} displays the results of these experiments.

\subsection{Baseline Agent Performance}
\label{sec:baselines}

As baselines, we evaluate our base agent model $\mathcal{M}$ as well as implement a trivial agent that always outputs \texttt{\small stop [N/A]}. A number of tasks in WebArena can be solved by this trivially implementable agent or a weak model that always refuses to continue and exits immediately, therefore, our trivial agent baseline helps discriminate which tasks being completed successfully should contribute to an agent being meaningfully capable when computing the capability score.

\subsection{Self-Improvement Fine-Tuned Agent Performance}

We fine-tune our base agent model $\mathcal{M}$ on the 3 synthetic dataset mixtures previously discussed: 1) $\mathcal{D}_A = \mathcal{D}_{\textsc{In-Domain}}$ 2) $\mathcal{D}_B = \mathcal{D}_{\textsc{In-Domain}} \cup \mathcal{D}_{\textsc{Out-of-Domain}}$ and 3) $\mathcal{D}_C = \mathcal{D}_{\textsc{Out-of-Domain}}$ with a straightforward auto-regressive loss using QLoRA \citep{qlora,lora}:
\[
L_{\text{FT}}(\theta) = -\mathbb{E}_{[(\mathbf{i}, o_t, a_{t-1}), a_t] \sim \mathcal{D}} \left[ \log P_\theta(a_t \mid (\mathbf{i}, o_t, a_{t-1})) \right]
\]

to produce $\mathcal{M}_{A}$, $\mathcal{M}_{B}$, and $\mathcal{M}_{C}$. We perform a 90/10\% train-validation split of our datasets and train with an early stopping patience of 5 epochs, using a batch size of 16 examples and a learning rate of 1e-5. Further details about training configuration and hyperparameters can be found in Appendix \ref{sec:appendix_training_and_inference_details}.

\subsection{Iterative Self-Improvement Fine-Tuned Agent Performance}

We also experiment with iterative self-improvement \citep{selfplay} to assess whether further improvement can be gained from a subsequent round of our self-improvement procedure. We perform this experiment on Mixture A. It is conceivable that after fine-tuning on $\mathcal{D}_A^1$, filtering from a set of trajectories with higher performance might yield a stronger set of plausible trajectories \footnote{To maximize data for iterative self-improvement, during filtering, we also fallback to checking the base model trajectory for a task if the self-improved model's trajectory for a task is filtered out.}  to produce $\mathcal{D}_A^2$. Mixtures B and C are less likely to demonstrate improvement over a subsequent round since the fine-tuned models are not specifically trained to generate better synthetic out-of-domain examples.
\experimentstable

\section{Discussion}
\label{sec:discussion}

We summarize key results from our experiments as well as discuss insights towards the efficacy of our self-improvement procedures for complex, multi-step tasks like web agent tasks.

\paragraph{Can models self-improve at web agent tasks?} We find fine-tuning on both Mixtures A and B improve overall benchmark performance with the best performing mixture, Mixture B, completing 18 more tasks correctly, a 31\% relative improvement ($7.14 \rightarrow 9.36$). Training on all Mixtures A, B, and C demonstrate self-improvement on at least one metric, with $\mathcal{M}_C$ showing a gain on capability score.

\paragraph{Do self-improved agents acquire new capabilities?}  We find agent models can acquire new capabilities through self-improvement, however, they also may lose the ability to perform some capabilities. In net, all of our self-improved agents acquire more capabilities than they lose. We find fine-tuning on both Mixtures A and B improve the capability score equally and lead to the largest net acquisition of capabilities demonstrating 5 more capabilities than the base agent model, a 24\% relative improvement ($15.44 \rightarrow 19.12$). We find all agent models demonstrate at least one new capability that no other agent model demonstrates, for example, only $\mathcal{M}_C$ successfully completes the ``Fork \texttt{\small \{\{repo\}\}}'' capability on the GitLab website. Interestingly, we find that the majority of capabilities acquired by $\mathcal{M}_A$ and $\mathcal{M}_C$ are mutually exclusive, suggesting in-domain synthetic examples and out-of-domain synthetic examples improve acquisition of different capabilities. We list all capabilities $\mathcal{M}_A$, $\mathcal{M}_B$, and $\mathcal{M}_C$ acquire and lose compared to $\mathcal{M}$ in Appendix \ref{sec:appendix_capabilityanalysis}.

\paragraph{Are self-improved agents more robust?} For $\mathcal{M}_B$, we find a larger improvement in functional correctness (31\%) than in capability score (24\%), which supports that the agent model is improving at more consistently succeeding at tasks belonging to the same capability, an indicator of one type of robustness. $\mathcal{M}_C$ is less robust by the same measure. Moreover, the capability analysis in Appendix \ref{sec:appendix_capabilityanalysis} also shows both $\mathcal{M}_A$ and $\mathcal{M}_B$ after self-improvement still demonstrate the majority of capabilities demonstrated by the base agent model $\mathcal{M}$, whereas $\mathcal{M}_C$ only demonstrates a minority. This would indicate $\mathcal{M}_A$ and $\mathcal{M}_B$ more reliably maintain the capabilities of the base agent model after self-improvement, a measure of robustness that would be useful in deployed settings where users of agent models may desire stability in performance.

\paragraph{Is there an effect on the quality of generated trajectories?} Fine-tuning on Mixtures A and B show no degradation in the quality of generated trajectories and show small improvement towards the reference on $\text{VERTEX}_{\text{DTW}}$. Fine-tuning on Mixture C degrades the the quality of generated trajectories from the reference. Training on the out-of-domain synthetic examples allows $\mathcal{M}_C$ to demonstrate some unique capabilities no other agent model demonstrates, however, inspecting trajectories from $\mathcal{M}_C$, we find this comes with trade-offs. For example, compared with $\mathcal{M}_A$, we find $\mathcal{M}_C$ produces longer trajectories (\textasciitilde 1.6x) and produces more invalid actions  (\textasciitilde 3.9x). In comparison with $\mathcal{M}$, $\mathcal{M}_A$ and  $\mathcal{M}_B$ do not greatly increase trajectory length (\textasciitilde 1.1x and \textasciitilde 1.3x) or the rate of invalid actions (\textasciitilde 1x and \textasciitilde 1.3x), further explaining the quality difference $\text{VERTEX}_{\text{DTW}}$ highlights. Due to lack of human reference, the reliability of this evaluation is limited which we discuss in Section \ref{sec:limitations}. In Appendix \ref{sec:appendix_fullvertexresults}, we compute variants of $\text{VERTEX}_{\text{DTW}}$, weighting by capability and filtering out trivial tasks. We find these variants make little difference in the relative ranking of agent models.

\paragraph{Can models iteratively self-improve at web agent tasks?} Our results are consistent with prior works such as \cite{selfplay} and \cite{Feng:24tale_of_tails} and we find diminishing returns to successive rounds of self-improvement and training on synthetic data. While the agent model after a second round of self-improvement outperforms the base agent model, it does not perform any better than agent models with a single round of self-improvement. We analyze the set of plausible trajectories in the second round in Appendix \ref{sec:appendix_iterativeselfimprovementplausibletrajectories} and find that while more synthetic training examples can be collected, they are of lower quality and contain a higher proportion of failed trajectories.

\section{Related Work}
\label{sec:relatedwork}

\paragraph{Self-Improvement} A number of techniques have been proposed for self-improving LLMs \citep[\textit{inter alia}]{llmscanselfimprove,selfverification,Madaan:23selfrefine}. Some self-improvement techniques \citep{unsupervisedmt,gulcehre2023reinforced,Singh:24rest_em,selfplay,selfrewarding} involve self-distillation \citep{selfdistill}, a special form of knowledge distillation \citep{Hinton:15distilling} where the teacher and student are the same model. A growing trend of works \citep{Wang:23selfinstruct,Gunasekar:23textbooks} similarly prompt LLMs to generate synthetic fine-tuning data.

\paragraph{LLM Agents} A number of prompting techniques proposed \citep{kojima2023large,Wei:22cot, react,shinn2023reflexion} can improve an LLM agent's performance, however, these techniques are orthogonal to self-improvement fine-tuning. \cite{chen2023fireact} introduces a technique for supervised fine-tuning of LLM agents. \cite{sodhi2024step} and \cite{lai2024autowebglm} introduce handcrafted subprompts or supervised techniques that improve performance on WebArena.

\paragraph{Self-Improving Agents} \cite{bousmalis2023robocat} demonstrates self-improving embodied agents for complex robotics tasks. \cite{Aksitov:24rest_react} introduces a method for self-improving agents on a simpler multi-step question answering task. Concurrently, \cite{trialanderror} proposes a similar procedure of filtering trajectories and fine-tuning, but primarily focuses on supervised filtering, does not explore generating novel tasks and synthetic data, and evaluates on less realistic and complex benchmarks. \cite{pan2024autonomous} explores using vision models for critique to improve on WebArena.

\section{Limitations and Broader Impacts}
\label{sec:limitations}

While we find self-improvement fine-tuning techniques can improve performance by reinforcing correct actions and decisions of an underlying model, these techniques can also further reinforce incorrect actions and biases of the underlying model. Some human or supervised filtering may mitigate this drawback, however, in this paper we focus our investigation on the efficacy and quality of unsupervised self-improvement as producing datasets for such complex tasks is difficult and expensive. Our analysis of capabilities is limited by our method to group tasks by the intent template used and cosine similarity. It is possible other strategies may produce more optimal groups to measure capabilities. Our  $\text{VERTEX}_{\text{DTW}}$ score utilizes a stronger model's generations (GPT-4) as a reference, however, human references would significantly improve the reliability of this evaluation. While WebArena spans many different types of realistic tasks and websites (shopping, online forums, maps, etc.), a future direction for this work might involve evaluation on larger, and more diverse benchmark.

\section{Conclusion}
\label{sec:conclusion}
In this work, we explore whether large language models can self-improve beyond their base performance at complex, long-horizon web agent tasks. We conclude self-improvement can increase the performance and robustness of agent models and allow agent models to acquire new capabilities. We also find it is possible for self-improvement to yield these benefits with minimal degradation to the quality of trajectories. The self-improvement procedures we propose are a promising step towards boosting the performance of LLMs in complex, multi-step agent environments such as web environments, without relying on supervised training data. We release our code, evaluation metrics with references, synthetic datasets, and model trajectories.
\section*{Acknowledgements}

This research is supported in part by the Office of the Director of National Intelligence (ODNI), Intelligence Advanced Research Projects Activity (IARPA), via the HIATUS Program contract \#2022-22072200005. The views and conclusions contained herein are those of the authors and should not be interpreted as necessarily representing the official policies, either expressed or implied, of ODNI, IARPA, or the U.S. Government. The U.S. Government is authorized to reproduce and distribute reprints for governmental purposes notwithstanding any copyright annotation therein.

The ELLIS Unit Linz, the LIT AI Lab, the Institute for Machine Learning, are supported by the Federal State Upper Austria. We thank the projects Medical Cognitive Computing Center (MC3), INCONTROL-RL (FFG-881064), PRIMAL (FFG-873979), S3AI (FFG-872172), DL for GranularFlow (FFG-871302), EPILEPSIA (FFG-892171), AIRI FG 9-N (FWF-36284, FWF-36235), AI4GreenHeatingGrids (FFG- 899943), INTEGRATE (FFG-892418), ELISE (H2020-ICT-2019-3 ID: 951847), Stars4Waters (HORIZON-CL6-2021-CLIMATE-01-01). We thank Audi.JKU Deep Learning Center, TGW LOGISTICS GROUP GMBH, Silicon Austria Labs (SAL), FILL Gesellschaft mbH, Anyline GmbH, Google, ZF Friedrichshafen AG, Robert Bosch GmbH, UCB Biopharma SRL, Merck Healthcare KGaA, Verbund AG, GLS (Univ. Waterloo), Software Competence Center Hagenberg GmbH, Borealis AG, T\"{U}V Austria, Frauscher Sensonic, TRUMPF, the NVIDIA Corporation and ExtensityAI.

The authors gratefully acknowledge the HPC RIVR consortium (www.hpc-rivr.si) and EuroHPC JU (eurohpc-ju.europa.eu) for funding this research by providing computing resources of the HPC system Vega at the Institute of Information Science (www.izum.si).

The authors gratefully acknowledge OpenAI for providing research credits and computing resources for this research.

\bibliographystyle{abbrvnat}
{
\small
\bibliography{references}

\begin{thebibliography}{51}
\providecommand{\natexlab}[1]{#1}
\providecommand{\url}[1]{\texttt{#1}}
\expandafter\ifx\csname urlstyle\endcsname\relax
  \providecommand{\doi}[1]{doi: #1}\else
  \providecommand{\doi}{doi: \begingroup \urlstyle{rm}\Url}\fi

\bibitem[Aksitov et~al.(2024)Aksitov, Miryoosefi, Li, Li, Babayan, Kopparapu, Fisher, Guo, Prakash, Srinivasan, Zaheer, Yu, and Kumar]{Aksitov:24rest_react}
R.~Aksitov, S.~Miryoosefi, Z.~Li, D.~Li, S.~Babayan, K.~Kopparapu, Z.~Fisher, R.~Guo, S.~Prakash, P.~Srinivasan, M.~Zaheer, F.~Yu, and S.~Kumar.
\newblock Re{ST} meets react: Self-improvement for multi-step reasoning {LLM} agent.
\newblock In \emph{ICLR 2024 Workshop on Large Language Model (LLM) Agents}, 2024.
\newblock URL \url{https://openreview.net/forum?id=7xknRLr7QE}.

\bibitem[Bai et~al.(2023)Bai, Bai, Chu, Cui, Dang, Deng, Fan, Ge, Han, Huang, Hui, Ji, Li, Lin, Lin, Liu, Liu, Lu, Lu, Ma, Men, Ren, Ren, Tan, Tan, Tu, Wang, Wang, Wang, Wu, Xu, Xu, Yang, Yang, Yang, Yang, Yao, Yu, Yuan, Yuan, Zhang, Zhang, Zhang, Zhang, Zhou, Zhou, Zhou, and Zhu]{qwen}
J.~Bai, S.~Bai, Y.~Chu, Z.~Cui, K.~Dang, X.~Deng, Y.~Fan, W.~Ge, Y.~Han, F.~Huang, B.~Hui, L.~Ji, M.~Li, J.~Lin, R.~Lin, D.~Liu, G.~Liu, C.~Lu, K.~Lu, J.~Ma, R.~Men, X.~Ren, X.~Ren, C.~Tan, S.~Tan, J.~Tu, P.~Wang, S.~Wang, W.~Wang, S.~Wu, B.~Xu, J.~Xu, A.~Yang, H.~Yang, J.~Yang, S.~Yang, Y.~Yao, B.~Yu, H.~Yuan, Z.~Yuan, J.~Zhang, X.~Zhang, Y.~Zhang, Z.~Zhang, C.~Zhou, J.~Zhou, X.~Zhou, and T.~Zhu.
\newblock Qwen technical report, 2023.

\bibitem[Berndt and Clifford(1994)]{Berndt:94dtw}
D.~J. Berndt and J.~Clifford.
\newblock Using dynamic time warping to find patterns in time series.
\newblock In \emph{KDD Workshop}, 1994.
\newblock URL \url{https://api.semanticscholar.org/CorpusID:929893}.

\bibitem[Bousmalis et~al.(2023)Bousmalis, Vezzani, Rao, Devin, Lee, Bauza, Davchev, Zhou, Gupta, Raju, Laurens, Fantacci, Dalibard, Zambelli, Martins, Pevceviciute, Blokzijl, Denil, Batchelor, Lampe, Parisotto, Żołna, Reed, Colmenarejo, Scholz, Abdolmaleki, Groth, Regli, Sushkov, Rothörl, Chen, Aytar, Barker, Ortiz, Riedmiller, Springenberg, Hadsell, Nori, and Heess]{bousmalis2023robocat}
K.~Bousmalis, G.~Vezzani, D.~Rao, C.~Devin, A.~X. Lee, M.~Bauza, T.~Davchev, Y.~Zhou, A.~Gupta, A.~Raju, A.~Laurens, C.~Fantacci, V.~Dalibard, M.~Zambelli, M.~Martins, R.~Pevceviciute, M.~Blokzijl, M.~Denil, N.~Batchelor, T.~Lampe, E.~Parisotto, K.~Żołna, S.~Reed, S.~G. Colmenarejo, J.~Scholz, A.~Abdolmaleki, O.~Groth, J.-B. Regli, O.~Sushkov, T.~Rothörl, J.~E. Chen, Y.~Aytar, D.~Barker, J.~Ortiz, M.~Riedmiller, J.~T. Springenberg, R.~Hadsell, F.~Nori, and N.~Heess.
\newblock Robocat: A self-improving generalist agent for robotic manipulation, 2023.

\bibitem[Brown et~al.(2020)Brown, Mann, Ryder, Subbiah, Kaplan, Dhariwal, Neelakantan, Shyam, Sastry, Askell, et~al.]{gpt3}
T.~Brown, B.~Mann, N.~Ryder, M.~Subbiah, J.~D. Kaplan, P.~Dhariwal, A.~Neelakantan, P.~Shyam, G.~Sastry, A.~Askell, et~al.
\newblock Language models are few-shot learners.
\newblock \emph{Advances in neural information processing systems}, 33:\penalty0 1877--1901, 2020.

\bibitem[Chen et~al.(2023)Chen, Shu, Shareghi, Collier, Narasimhan, and Yao]{chen2023fireact}
B.~Chen, C.~Shu, E.~Shareghi, N.~Collier, K.~Narasimhan, and S.~Yao.
\newblock Fireact: Toward language agent fine-tuning, 2023.

\bibitem[Chen et~al.(2024)Chen, Deng, Yuan, Ji, and Gu]{selfplay}
Z.~Chen, Y.~Deng, H.~Yuan, K.~Ji, and Q.~Gu.
\newblock Self-play fine-tuning converts weak language models to strong language models, 2024.

\bibitem[Chiang et~al.(2024)Chiang, Zheng, Sheng, Angelopoulos, Li, Li, Zhang, Zhu, Jordan, Gonzalez, and Stoica]{lmysys}
W.-L. Chiang, L.~Zheng, Y.~Sheng, A.~N. Angelopoulos, T.~Li, D.~Li, H.~Zhang, B.~Zhu, M.~Jordan, J.~E. Gonzalez, and I.~Stoica.
\newblock Chatbot arena: An open platform for evaluating llms by human preference, 2024.

\bibitem[Dettmers et~al.(2023)Dettmers, Pagnoni, Holtzman, and Zettlemoyer]{qlora}
T.~Dettmers, A.~Pagnoni, A.~Holtzman, and L.~Zettlemoyer.
\newblock Qlora: Efficient finetuning of quantized llms, 2023.

\bibitem[Dinu et~al.(2024)Dinu, Leoveanu-Condrei, Holzleitner, Zellinger, and Hochreiter]{symbolicai}
M.-C. Dinu, C.~Leoveanu-Condrei, M.~Holzleitner, W.~Zellinger, and S.~Hochreiter.
\newblock Symbolicai: A framework for logic-based approaches combining generative models and solvers, 2024.

\bibitem[Feng et~al.(2024)Feng, Dohmatob, Yang, Charton, and Kempe]{Feng:24tale_of_tails}
Y.~Feng, E.~Dohmatob, P.~Yang, F.~Charton, and J.~Kempe.
\newblock A tale of tails: Model collapse as a change of scaling laws.
\newblock 2024.
\newblock URL \url{https://openreview.net/forum?id=dE8BznbvZV}.

\bibitem[Gou et~al.(2024)Gou, Shao, Gong, Shen, Yang, Duan, and Chen]{environmentcritique}
Z.~Gou, Z.~Shao, Y.~Gong, Y.~Shen, Y.~Yang, N.~Duan, and W.~Chen.
\newblock Critic: Large language models can self-correct with tool-interactive critiquing, 2024.

\bibitem[Gulcehre et~al.(2023)Gulcehre, Paine, Srinivasan, Konyushkova, Weerts, Sharma, Siddhant, Ahern, Wang, Gu, Macherey, Doucet, Firat, and de~Freitas]{gulcehre2023reinforced}
C.~Gulcehre, T.~L. Paine, S.~Srinivasan, K.~Konyushkova, L.~Weerts, A.~Sharma, A.~Siddhant, A.~Ahern, M.~Wang, C.~Gu, W.~Macherey, A.~Doucet, O.~Firat, and N.~de~Freitas.
\newblock Reinforced self-training (rest) for language modeling, 2023.

\bibitem[Gunasekar et~al.(2023)Gunasekar, Zhang, Aneja, Mendes, Giorno, Gopi, Javaheripi, Kauffmann, de~Rosa, Saarikivi, Salim, Shah, Behl, Wang, Bubeck, Eldan, Kalai, Lee, and Li]{Gunasekar:23textbooks}
S.~Gunasekar, Y.~Zhang, J.~Aneja, C.~C.~T. Mendes, A.~D. Giorno, S.~Gopi, M.~Javaheripi, P.~Kauffmann, G.~de~Rosa, O.~Saarikivi, A.~Salim, S.~Shah, H.~S. Behl, X.~Wang, S.~Bubeck, R.~Eldan, A.~T. Kalai, Y.~T. Lee, and Y.~Li.
\newblock Textbooks are all you need.
\newblock 2023.

\bibitem[Han et~al.(2021)Han, Babuschkin, Edwards, Neelakantan, Xu, Polu, Ray, Shyam, Ramesh, Radford, and Sutskever]{unsupervisedmt}
J.~M. Han, I.~Babuschkin, H.~Edwards, A.~Neelakantan, T.~Xu, S.~Polu, A.~Ray, P.~Shyam, A.~Ramesh, A.~Radford, and I.~Sutskever.
\newblock Unsupervised neural machine translation with generative language models only.
\newblock \emph{CoRR}, abs/2110.05448, 2021.
\newblock URL \url{https://arxiv.org/abs/2110.05448}.

\bibitem[Hinton et~al.(2015)Hinton, Vinyals, and Dean]{Hinton:15distilling}
G.~Hinton, O.~Vinyals, and J.~Dean.
\newblock Distilling the knowledge in a neural network.
\newblock 2015.

\bibitem[Hu et~al.(2021)Hu, Shen, Wallis, Allen-Zhu, Li, Wang, Wang, and Chen]{lora}
E.~J. Hu, Y.~Shen, P.~Wallis, Z.~Allen-Zhu, Y.~Li, S.~Wang, L.~Wang, and W.~Chen.
\newblock Lora: Low-rank adaptation of large language models, 2021.

\bibitem[Huang et~al.(2022)Huang, Gu, Hou, Wu, Wang, Yu, and Han]{llmscanselfimprove}
J.~Huang, S.~S. Gu, L.~Hou, Y.~Wu, X.~Wang, H.~Yu, and J.~Han.
\newblock Large language models can self-improve, 2022.

\bibitem[Kojima et~al.(2023)Kojima, Gu, Reid, Matsuo, and Iwasawa]{kojima2023large}
T.~Kojima, S.~S. Gu, M.~Reid, Y.~Matsuo, and Y.~Iwasawa.
\newblock Large language models are zero-shot reasoners, 2023.

\bibitem[Lai et~al.(2024)Lai, Liu, Iong, Yao, Chen, Shen, Yu, Zhang, Zhang, Dong, and Tang]{lai2024autowebglm}
H.~Lai, X.~Liu, I.~L. Iong, S.~Yao, Y.~Chen, P.~Shen, H.~Yu, H.~Zhang, X.~Zhang, Y.~Dong, and J.~Tang.
\newblock Autowebglm: Bootstrap and reinforce a large language model-based web navigating agent, 2024.

\bibitem[Liu et~al.(2019)Liu, Ott, Goyal, Du, Joshi, Chen, Levy, Lewis, Zettlemoyer, and Stoyanov]{roberta}
Y.~Liu, M.~Ott, N.~Goyal, J.~Du, M.~Joshi, D.~Chen, O.~Levy, M.~Lewis, L.~Zettlemoyer, and V.~Stoyanov.
\newblock Roberta: {A} robustly optimized {BERT} pretraining approach.
\newblock \emph{CoRR}, abs/1907.11692, 2019.
\newblock URL \url{http://arxiv.org/abs/1907.11692}.

\bibitem[Madaan et~al.(2023)Madaan, Tandon, Gupta, Hallinan, Gao, Wiegreffe, Alon, Dziri, Prabhumoye, Yang, Gupta, Majumder, Hermann, Welleck, Yazdanbakhsh, and Clark]{Madaan:23selfrefine}
A.~Madaan, N.~Tandon, P.~Gupta, S.~Hallinan, L.~Gao, S.~Wiegreffe, U.~Alon, N.~Dziri, S.~Prabhumoye, Y.~Yang, S.~Gupta, B.~P. Majumder, K.~Hermann, S.~Welleck, A.~Yazdanbakhsh, and P.~Clark.
\newblock Self-refine: Iterative refinement with self-feedback.
\newblock In \emph{Thirty-seventh Conference on Neural Information Processing Systems}, 2023.
\newblock URL \url{https://openreview.net/forum?id=S37hOerQLB}.

\bibitem[OpenAI et~al.(2024)OpenAI, Achiam, Adler, Agarwal, Ahmad, Akkaya, Aleman, Almeida, Altenschmidt, Altman, Anadkat, Avila, Babuschkin, Balaji, Balcom, Baltescu, Bao, Bavarian, Belgum, Bello, Berdine, Bernadett-Shapiro, Berner, Bogdonoff, Boiko, Boyd, Brakman, Brockman, Brooks, Brundage, Button, Cai, Campbell, Cann, Carey, Carlson, Carmichael, Chan, Chang, Chantzis, Chen, Chen, Chen, Chen, Chen, Chess, Cho, Chu, Chung, Cummings, Currier, Dai, Decareaux, Degry, Deutsch, Deville, Dhar, Dohan, Dowling, Dunning, Ecoffet, Eleti, Eloundou, Farhi, Fedus, Felix, Fishman, Forte, Fulford, Gao, Georges, Gibson, Goel, Gogineni, Goh, Gontijo-Lopes, Gordon, Grafstein, Gray, Greene, Gross, Gu, Guo, Hallacy, Han, Harris, He, Heaton, Heidecke, Hesse, Hickey, Hickey, Hoeschele, Houghton, Hsu, Hu, Hu, Huizinga, Jain, Jain, Jang, Jiang, Jiang, Jin, Jin, Jomoto, Jonn, Jun, Kaftan, Łukasz Kaiser, Kamali, Kanitscheider, Keskar, Khan, Kilpatrick, Kim, Kim, Kim, Kirchner, Kiros, Knight, Kokotajlo, Łukasz Kondraciuk, Kondrich,
  Konstantinidis, Kosic, Krueger, Kuo, Lampe, Lan, Lee, Leike, Leung, Levy, Li, Lim, Lin, Lin, Litwin, Lopez, Lowe, Lue, Makanju, Malfacini, Manning, Markov, Markovski, Martin, Mayer, Mayne, McGrew, McKinney, McLeavey, McMillan, McNeil, Medina, Mehta, Menick, Metz, Mishchenko, Mishkin, Monaco, Morikawa, Mossing, Mu, Murati, Murk, Mély, Nair, Nakano, Nayak, Neelakantan, Ngo, Noh, Ouyang, O'Keefe, Pachocki, Paino, Palermo, Pantuliano, Parascandolo, Parish, Parparita, Passos, Pavlov, Peng, Perelman, de~Avila Belbute~Peres, Petrov, de~Oliveira~Pinto, Michael, Pokorny, Pokrass, Pong, Powell, Power, Power, Proehl, Puri, Radford, Rae, Ramesh, Raymond, Real, Rimbach, Ross, Rotsted, Roussez, Ryder, Saltarelli, Sanders, Santurkar, Sastry, Schmidt, Schnurr, Schulman, Selsam, Sheppard, Sherbakov, Shieh, Shoker, Shyam, Sidor, Sigler, Simens, Sitkin, Slama, Sohl, Sokolowsky, Song, Staudacher, Such, Summers, Sutskever, Tang, Tezak, Thompson, Tillet, Tootoonchian, Tseng, Tuggle, Turley, Tworek, Uribe, Vallone, Vijayvergiya,
  Voss, Wainwright, Wang, Wang, Wang, Ward, Wei, Weinmann, Welihinda, Welinder, Weng, Weng, Wiethoff, Willner, Winter, Wolrich, Wong, Workman, Wu, Wu, Wu, Xiao, Xu, Yoo, Yu, Yuan, Zaremba, Zellers, Zhang, Zhang, Zhao, Zheng, Zhuang, Zhuk, and Zoph]{gpt4}
OpenAI, J.~Achiam, S.~Adler, S.~Agarwal, L.~Ahmad, I.~Akkaya, F.~L. Aleman, D.~Almeida, J.~Altenschmidt, S.~Altman, S.~Anadkat, R.~Avila, I.~Babuschkin, S.~Balaji, V.~Balcom, P.~Baltescu, H.~Bao, M.~Bavarian, J.~Belgum, I.~Bello, J.~Berdine, G.~Bernadett-Shapiro, C.~Berner, L.~Bogdonoff, O.~Boiko, M.~Boyd, A.-L. Brakman, G.~Brockman, T.~Brooks, M.~Brundage, K.~Button, T.~Cai, R.~Campbell, A.~Cann, B.~Carey, C.~Carlson, R.~Carmichael, B.~Chan, C.~Chang, F.~Chantzis, D.~Chen, S.~Chen, R.~Chen, J.~Chen, M.~Chen, B.~Chess, C.~Cho, C.~Chu, H.~W. Chung, D.~Cummings, J.~Currier, Y.~Dai, C.~Decareaux, T.~Degry, N.~Deutsch, D.~Deville, A.~Dhar, D.~Dohan, S.~Dowling, S.~Dunning, A.~Ecoffet, A.~Eleti, T.~Eloundou, D.~Farhi, L.~Fedus, N.~Felix, S.~P. Fishman, J.~Forte, I.~Fulford, L.~Gao, E.~Georges, C.~Gibson, V.~Goel, T.~Gogineni, G.~Goh, R.~Gontijo-Lopes, J.~Gordon, M.~Grafstein, S.~Gray, R.~Greene, J.~Gross, S.~S. Gu, Y.~Guo, C.~Hallacy, J.~Han, J.~Harris, Y.~He, M.~Heaton, J.~Heidecke, C.~Hesse, A.~Hickey,
  W.~Hickey, P.~Hoeschele, B.~Houghton, K.~Hsu, S.~Hu, X.~Hu, J.~Huizinga, S.~Jain, S.~Jain, J.~Jang, A.~Jiang, R.~Jiang, H.~Jin, D.~Jin, S.~Jomoto, B.~Jonn, H.~Jun, T.~Kaftan, Łukasz Kaiser, A.~Kamali, I.~Kanitscheider, N.~S. Keskar, T.~Khan, L.~Kilpatrick, J.~W. Kim, C.~Kim, Y.~Kim, J.~H. Kirchner, J.~Kiros, M.~Knight, D.~Kokotajlo, Łukasz Kondraciuk, A.~Kondrich, A.~Konstantinidis, K.~Kosic, G.~Krueger, V.~Kuo, M.~Lampe, I.~Lan, T.~Lee, J.~Leike, J.~Leung, D.~Levy, C.~M. Li, R.~Lim, M.~Lin, S.~Lin, M.~Litwin, T.~Lopez, R.~Lowe, P.~Lue, A.~Makanju, K.~Malfacini, S.~Manning, T.~Markov, Y.~Markovski, B.~Martin, K.~Mayer, A.~Mayne, B.~McGrew, S.~M. McKinney, C.~McLeavey, P.~McMillan, J.~McNeil, D.~Medina, A.~Mehta, J.~Menick, L.~Metz, A.~Mishchenko, P.~Mishkin, V.~Monaco, E.~Morikawa, D.~Mossing, T.~Mu, M.~Murati, O.~Murk, D.~Mély, A.~Nair, R.~Nakano, R.~Nayak, A.~Neelakantan, R.~Ngo, H.~Noh, L.~Ouyang, C.~O'Keefe, J.~Pachocki, A.~Paino, J.~Palermo, A.~Pantuliano, G.~Parascandolo, J.~Parish, E.~Parparita,
  A.~Passos, M.~Pavlov, A.~Peng, A.~Perelman, F.~de~Avila Belbute~Peres, M.~Petrov, H.~P. de~Oliveira~Pinto, Michael, Pokorny, M.~Pokrass, V.~H. Pong, T.~Powell, A.~Power, B.~Power, E.~Proehl, R.~Puri, A.~Radford, J.~Rae, A.~Ramesh, C.~Raymond, F.~Real, K.~Rimbach, C.~Ross, B.~Rotsted, H.~Roussez, N.~Ryder, M.~Saltarelli, T.~Sanders, S.~Santurkar, G.~Sastry, H.~Schmidt, D.~Schnurr, J.~Schulman, D.~Selsam, K.~Sheppard, T.~Sherbakov, J.~Shieh, S.~Shoker, P.~Shyam, S.~Sidor, E.~Sigler, M.~Simens, J.~Sitkin, K.~Slama, I.~Sohl, B.~Sokolowsky, Y.~Song, N.~Staudacher, F.~P. Such, N.~Summers, I.~Sutskever, J.~Tang, N.~Tezak, M.~B. Thompson, P.~Tillet, A.~Tootoonchian, E.~Tseng, P.~Tuggle, N.~Turley, J.~Tworek, J.~F.~C. Uribe, A.~Vallone, A.~Vijayvergiya, C.~Voss, C.~Wainwright, J.~J. Wang, A.~Wang, B.~Wang, J.~Ward, J.~Wei, C.~Weinmann, A.~Welihinda, P.~Welinder, J.~Weng, L.~Weng, M.~Wiethoff, D.~Willner, C.~Winter, S.~Wolrich, H.~Wong, L.~Workman, S.~Wu, J.~Wu, M.~Wu, K.~Xiao, T.~Xu, S.~Yoo, K.~Yu, Q.~Yuan,
  W.~Zaremba, R.~Zellers, C.~Zhang, M.~Zhang, S.~Zhao, T.~Zheng, J.~Zhuang, W.~Zhuk, and B.~Zoph.
\newblock Gpt-4 technical report, 2024.

\bibitem[Ouyang et~al.(2022)Ouyang, Wu, Jiang, Almeida, Wainwright, Mishkin, Zhang, Agarwal, Slama, Ray, et~al.]{instructgpt}
L.~Ouyang, J.~Wu, X.~Jiang, D.~Almeida, C.~Wainwright, P.~Mishkin, C.~Zhang, S.~Agarwal, K.~Slama, A.~Ray, et~al.
\newblock Training language models to follow instructions with human feedback.
\newblock \emph{Advances in Neural Information Processing Systems}, 35:\penalty0 27730--27744, 2022.

\bibitem[Pan et~al.(2024)Pan, Zhang, Tomlin, Zhou, Levine, and Suhr]{pan2024autonomous}
J.~Pan, Y.~Zhang, N.~Tomlin, Y.~Zhou, S.~Levine, and A.~Suhr.
\newblock Autonomous evaluation and refinement of digital agents, 2024.

\bibitem[Paszke et~al.(2019)Paszke, Gross, Massa, Lerer, Bradbury, Chanan, Killeen, Lin, Gimelshein, Antiga, Desmaison, K{\"{o}}pf, Yang, DeVito, Raison, Tejani, Chilamkurthy, Steiner, Fang, Bai, and Chintala]{pytorch}
A.~Paszke, S.~Gross, F.~Massa, A.~Lerer, J.~Bradbury, G.~Chanan, T.~Killeen, Z.~Lin, N.~Gimelshein, L.~Antiga, A.~Desmaison, A.~K{\"{o}}pf, E.~Z. Yang, Z.~DeVito, M.~Raison, A.~Tejani, S.~Chilamkurthy, B.~Steiner, L.~Fang, J.~Bai, and S.~Chintala.
\newblock Pytorch: An imperative style, high-performance deep learning library.
\newblock \emph{CoRR}, abs/1912.01703, 2019.
\newblock URL \url{http://arxiv.org/abs/1912.01703}.

\bibitem[Patel et~al.(2023)Patel, Li, Rasooli, Constant, Raffel, and Callison-Burch]{bootstrapmtexamples}
A.~Patel, B.~Li, M.~S. Rasooli, N.~Constant, C.~Raffel, and C.~Callison-Burch.
\newblock Bidirectional language models are also few-shot learners, 2023.

\bibitem[Patel et~al.(2024)Patel, Raffel, and Callison-Burch]{datadreamer}
A.~Patel, C.~Raffel, and C.~Callison-Burch.
\newblock Datadreamer: A tool for synthetic data generation and reproducible llm workflows, 2024.

\bibitem[Pham et~al.(2022)Pham, Cho, Joshi, and Hegde]{selfdistill2}
M.~Pham, M.~Cho, A.~Joshi, and C.~Hegde.
\newblock Revisiting self-distillation, 2022.

\bibitem[Radford et~al.(2019)Radford, Wu, Child, Luan, Amodei, and Sutskever]{gpt2}
A.~Radford, J.~Wu, R.~Child, D.~Luan, D.~Amodei, and I.~Sutskever.
\newblock Language models are unsupervised multitask learners.
\newblock 2019.

\bibitem[Raffel et~al.(2020)Raffel, Shazeer, Roberts, Lee, Narang, Matena, Zhou, Li, and Liu]{t5}
C.~Raffel, N.~Shazeer, A.~Roberts, K.~Lee, S.~Narang, M.~Matena, Y.~Zhou, W.~Li, and P.~J. Liu.
\newblock Exploring the limits of transfer learning with a unified text-to-text transformer.
\newblock \emph{The Journal of Machine Learning Research}, 21\penalty0 (1):\penalty0 5485--5551, 2020.

\bibitem[Reimers and Gurevych(2019)]{sentencetransformers}
N.~Reimers and I.~Gurevych.
\newblock Sentence-bert: Sentence embeddings using siamese bert-networks.
\newblock In \emph{Proceedings of the 2019 Conference on Empirical Methods in Natural Language Processing and the 9th International Joint Conference on Natural Language Processing (EMNLP-IJCNLP)}, pages 3982--3992, 2019.

\bibitem[Salvador and Chan(2004)]{Stan:04fastdtw}
S.~Salvador and P.~Chan.
\newblock Toward accurate dynamic time warping in linear time and space.
\newblock volume~11, pages 70--80, 01 2004.

\bibitem[Sanh et~al.(2019)Sanh, Debut, Chaumond, and Wolf]{distilroberta}
V.~Sanh, L.~Debut, J.~Chaumond, and T.~Wolf.
\newblock Distilbert, a distilled version of {BERT:} smaller, faster, cheaper and lighter.
\newblock \emph{CoRR}, abs/1910.01108, 2019.
\newblock URL \url{http://arxiv.org/abs/1910.01108}.

\bibitem[Shinn et~al.(2023)Shinn, Cassano, Berman, Gopinath, Narasimhan, and Yao]{shinn2023reflexion}
N.~Shinn, F.~Cassano, E.~Berman, A.~Gopinath, K.~Narasimhan, and S.~Yao.
\newblock Reflexion: Language agents with verbal reinforcement learning, 2023.

\bibitem[Singh et~al.(2024)Singh, Co-Reyes, Agarwal, Anand, Patil, Garcia, Liu, Harrison, Lee, Xu, Parisi, Kumar, Alemi, Rizkowsky, Nova, Adlam, Bohnet, Elsayed, Sedghi, Mordatch, Simpson, Gur, Snoek, Pennington, Hron, Kenealy, Swersky, Mahajan, Culp, Xiao, Bileschi, Constant, Novak, Liu, Warkentin, Bansal, Dyer, Neyshabur, Sohl-Dickstein, and Fiedel]{Singh:24rest_em}
A.~Singh, J.~D. Co-Reyes, R.~Agarwal, A.~Anand, P.~Patil, X.~Garcia, P.~J. Liu, J.~Harrison, J.~Lee, K.~Xu, A.~T. Parisi, A.~Kumar, A.~A. Alemi, A.~Rizkowsky, A.~Nova, B.~Adlam, B.~Bohnet, G.~F. Elsayed, H.~Sedghi, I.~Mordatch, I.~Simpson, I.~Gur, J.~Snoek, J.~Pennington, J.~Hron, K.~Kenealy, K.~Swersky, K.~Mahajan, L.~A. Culp, L.~Xiao, M.~Bileschi, N.~Constant, R.~Novak, R.~Liu, T.~Warkentin, Y.~Bansal, E.~Dyer, B.~Neyshabur, J.~Sohl-Dickstein, and N.~Fiedel.
\newblock Beyond human data: Scaling self-training for problem-solving with language models.
\newblock \emph{Transactions on Machine Learning Research}, 2024.
\newblock ISSN 2835-8856.
\newblock URL \url{https://openreview.net/forum?id=lNAyUngGFK}.
\newblock Expert Certification.

\bibitem[slaypni(2017)]{fastdtw:software}
slaypni.
\newblock fastdtw: Fast implementation of the dynamic time warping algorithm, 2017.
\newblock URL \url{https://github.com/slaypni/fastdtw}.
\newblock Accessed: 2024-05-22.

\bibitem[Sodhi et~al.(2024)Sodhi, Branavan, Artzi, and McDonald]{sodhi2024step}
P.~Sodhi, S.~R.~K. Branavan, Y.~Artzi, and R.~McDonald.
\newblock Step: Stacked llm policies for web actions, 2024.

\bibitem[Song et~al.(2020)Song, Tan, Qin, Lu, and Liu]{Song:20mpnet}
K.~Song, X.~Tan, T.~Qin, J.~Lu, and T.-Y. Liu.
\newblock Mpnet: Masked and permuted pre-training for language understanding.
\newblock \emph{arXiv preprint arXiv:2004.09297}, 2020.

\bibitem[Song et~al.(2024)Song, Yin, Yue, Huang, Li, and Lin]{trialanderror}
Y.~Song, D.~Yin, X.~Yue, J.~Huang, S.~Li, and B.~Y. Lin.
\newblock Trial and error: Exploration-based trajectory optimization for llm agents, 2024.

\bibitem[Wang et~al.(2023)Wang, Kordi, Mishra, Liu, Smith, Khashabi, and Hajishirzi]{Wang:23selfinstruct}
Y.~Wang, Y.~Kordi, S.~Mishra, A.~Liu, N.~A. Smith, D.~Khashabi, and H.~Hajishirzi.
\newblock Self-instruct: Aligning language models with self-generated instructions.
\newblock 2023.

\bibitem[Wei et~al.(2021)Wei, Bosma, Zhao, Guu, Yu, Lester, Du, Dai, and Le]{flan}
J.~Wei, M.~Bosma, V.~Y. Zhao, K.~Guu, A.~W. Yu, B.~Lester, N.~Du, A.~M. Dai, and Q.~V. Le.
\newblock Finetuned language models are zero-shot learners.
\newblock \emph{arXiv preprint arXiv:2109.01652}, 2021.

\bibitem[Wei et~al.(2022)Wei, Wang, Schuurmans, Bosma, Ichter, Xia, Chi, Le, and Zhou]{Wei:22cot}
J.~Wei, X.~Wang, D.~Schuurmans, M.~Bosma, B.~Ichter, F.~Xia, E.~H. Chi, Q.~V. Le, and D.~Zhou.
\newblock Chain-of-thought prompting elicits reasoning in large language models.
\newblock In \emph{NeurIPS}, 2022.
\newblock URL \url{http://papers.nips.cc/paper_files/paper/2022/hash/9d5609613524ecf4f15af0f7b31abca4-Abstract-Conference.html}.

\bibitem[Weng et~al.(2022)Weng, Zhu, Xia, Li, He, Liu, Sun, Liu, and Zhao]{selfverification}
Y.~Weng, M.~Zhu, F.~Xia, B.~Li, S.~He, S.~Liu, B.~Sun, K.~Liu, and J.~Zhao.
\newblock Large language models are better reasoners with self-verification.
\newblock \emph{arXiv preprint arXiv:2212.09561}, 2022.

\bibitem[Wolf et~al.(2019)Wolf, Debut, Sanh, Chaumond, Delangue, Moi, Cistac, Rault, Louf, Funtowicz, and Brew]{transformers}
T.~Wolf, L.~Debut, V.~Sanh, J.~Chaumond, C.~Delangue, A.~Moi, P.~Cistac, T.~Rault, R.~Louf, M.~Funtowicz, and J.~Brew.
\newblock Huggingface's transformers: State-of-the-art natural language processing.
\newblock \emph{CoRR}, abs/1910.03771, 2019.
\newblock URL \url{http://arxiv.org/abs/1910.03771}.

\bibitem[Yao et~al.(2023)Yao, Zhao, Yu, Du, Shafran, Narasimhan, and Cao]{react}
S.~Yao, J.~Zhao, D.~Yu, N.~Du, I.~Shafran, K.~Narasimhan, and Y.~Cao.
\newblock React: Synergizing reasoning and acting in language models, 2023.

\bibitem[Yuan et~al.(2024)Yuan, Pang, Cho, Li, Sukhbaatar, Xu, and Weston]{selfrewarding}
W.~Yuan, R.~Y. Pang, K.~Cho, X.~Li, S.~Sukhbaatar, J.~Xu, and J.~Weston.
\newblock Self-rewarding language models, 2024.

\bibitem[Yuan et~al.(2023)Yuan, Yuan, Li, Dong, Lu, Tan, Zhou, and Zhou]{rft}
Z.~Yuan, H.~Yuan, C.~Li, G.~Dong, K.~Lu, C.~Tan, C.~Zhou, and J.~Zhou.
\newblock Scaling relationship on learning mathematical reasoning with large language models, 2023.

\bibitem[Zhang et~al.(2019)Zhang, Song, Gao, Chen, Bao, and Ma]{selfdistill}
L.~Zhang, J.~Song, A.~Gao, J.~Chen, C.~Bao, and K.~Ma.
\newblock Be your own teacher: Improve the performance of convolutional neural networks via self distillation, 2019.

\bibitem[Zhao et~al.(2023)Zhao, Gu, Varma, Luo, Huang, Xu, Wright, Shojanazeri, Ott, Shleifer, et~al.]{fsdp}
Y.~Zhao, A.~Gu, R.~Varma, L.~Luo, C.-C. Huang, M.~Xu, L.~Wright, H.~Shojanazeri, M.~Ott, S.~Shleifer, et~al.
\newblock Pytorch fsdp: experiences on scaling fully sharded data parallel.
\newblock \emph{arXiv preprint arXiv:2304.11277}, 2023.

\bibitem[Zhou et~al.(2023)Zhou, Xu, Zhu, Zhou, Lo, Sridhar, Cheng, Bisk, Fried, Alon, et~al.]{webarena}
S.~Zhou, F.~F. Xu, H.~Zhu, X.~Zhou, R.~Lo, A.~Sridhar, X.~Cheng, Y.~Bisk, D.~Fried, U.~Alon, et~al.
\newblock Webarena: A realistic web environment for building autonomous agents.
\newblock \emph{arXiv preprint arXiv:2307.13854}, 2023.

\end{thebibliography}
}


\newpage\appendix
\section*{Appendix}
\addcontentsline{toc}{section}{Appendix}
\section{Training and Inference Details}
\label{sec:appendix_training_and_inference_details}

\hyperparamtable

\inferenceparamtable
\newpage
\section{Sample of Generated Out-of-Domain Synthetic Training Example}
\label{sec:full_outofdomain_example}

\fulloutofdomainexampletable
\newpage
\section{Capability Analysis}
\label{sec:appendix_capabilityanalysis}
\capabilityanalysistable
\newpage
\section{Full $\text{VERTEX}_{\text{DTW}}$ Score Results}
\label{sec:appendix_fullvertexresults}

\fullvertextable
\newpage
\section{Iterative Self-Improvement Plausible Trajectories}
\label{sec:appendix_iterativeselfimprovementplausibletrajectories}

\iterativefilteringtable
\newpage
\section{Capabilities in WebArena}
\label{sec:appendix_capabilities}

In this appendix, we list the grouping of tasks into ``capabilities'' we find in WebArena using the automated method we describe in Section \ref{sec:capabilityscore}. These tasks are grouped by the intent template used by WebArena to create the task as well as cosine similarity to group paraphrases detected by a sentence similarity model. We do not perform manual modifications to the groups and instead solely rely on automated techniques. We acknowledge grouping of natural language task objectives into capability areas is subjective and discuss this a limitation in Section \ref{sec:limitations}:
{
\scriptsize
\begin{longtable}[H!]
{p{.30\textwidth}p{.30\textwidth}p{.30\textwidth}}
{
\textbf{Capability \#1:}
\begin{itemize}[noitemsep,leftmargin=*]
   \item What are the top-\texttt{\tiny \{\{n\}\}} best-selling product in \texttt{\tiny \{\{year\}\}}
\end{itemize}
\textbf{Capability \#2:}
\begin{itemize}[noitemsep,leftmargin=*]
   \item Tell me the the number of reviews that our store received by far that mention term ``\texttt{\tiny \{\{term\}\}}''
\end{itemize}
\textbf{Capability \#3:}
\begin{itemize}[noitemsep,leftmargin=*]
   \item What brands appear most frequently among the top search terms?
   \item List the top \texttt{\tiny \{\{n\}\}} search terms in my store
\end{itemize}
\textbf{Capability \#4:}
\begin{itemize}[noitemsep,leftmargin=*]
   \item Telll me the grand total of invoice \texttt{\tiny \{\{id\}\}}.
\end{itemize}
\textbf{Capability \#5:}
\begin{itemize}[noitemsep,leftmargin=*]
   \item Presents the monthly count of successful orders \texttt{\tiny \{\{period\}\}} in MM:COUNT format
\end{itemize}
\textbf{Capability \#6:}
\begin{itemize}[noitemsep,leftmargin=*]
   \item What's the total number of items sold in the most recent \texttt{\tiny \{\{k\}\}} orders?
\end{itemize}
\textbf{Capability \#7:}
\begin{itemize}[noitemsep,leftmargin=*]
   \item Show all customers
\end{itemize}
\textbf{Capability \#8:}
\begin{itemize}[noitemsep,leftmargin=*]
   \item Give me the \texttt{\tiny \{\{Attribute\}\}} of the products that have \texttt{\tiny \{\{N\}\}} units left
\end{itemize}
\textbf{Capability \#9:}
\begin{itemize}[noitemsep,leftmargin=*]
   \item Get the total payment amount of the last \texttt{\tiny \{\{N\}\}} \texttt{\tiny \{\{status\}\}} orders
\end{itemize}
\textbf{Capability \#10:}
\begin{itemize}[noitemsep,leftmargin=*]
   \item Find the customer name and email with phone number \texttt{\tiny \{\{PhoneNum\}\}}
\end{itemize}
\textbf{Capability \#11:}
\begin{itemize}[noitemsep,leftmargin=*]
   \item Tell me the \texttt{\tiny \{\{attribute\}\}} of the customer who has the most cancellations in the history
   \item Which customer has completed the \texttt{\tiny \{\{quantifier\}\}} number of orders in the entire history?
   \item Show me the \texttt{\tiny \{\{information\}\}} of the customer who is the most unhappy with \texttt{\tiny \{\{product\}\}}
\end{itemize}
\textbf{Capability \#12:}
\begin{itemize}[noitemsep,leftmargin=*]
   \item How many reviews our shop received \texttt{\tiny \{\{time\}\}}?
   \item What is the total count of \texttt{\tiny \{\{status\}\}} reviews amongst all the reviews?
\end{itemize}
\textbf{Capability \#13:}
\begin{itemize}[noitemsep,leftmargin=*]
   \item Preview the \texttt{\tiny \{\{name\}\}} theme for my shop
\end{itemize}
\textbf{Capability \#14:}
\begin{itemize}[noitemsep,leftmargin=*]
   \item Mark all \texttt{\tiny \{\{brand\}\}} shirts on sale
\end{itemize}
} & {
\textbf{Capability \#15:}
\begin{itemize}[noitemsep,leftmargin=*]
   \item Disable \texttt{\tiny \{\{product\}\}} from the site, they are facing some quality issues.
\end{itemize}
\textbf{Capability \#16:}
\begin{itemize}[noitemsep,leftmargin=*]
   \item \texttt{\tiny \{\{action\}\}} the price of \texttt{\tiny \{\{config\}\}} by \texttt{\tiny \{\{amount\}\}}
   \item \texttt{\tiny \{\{action\}\}} the price of this product by \texttt{\tiny \{\{amount\}\}}
\end{itemize}
\textbf{Capability \#17:}
\begin{itemize}[noitemsep,leftmargin=*]
   \item Update the description of \texttt{\tiny \{\{product\}\}} to highlight the real user positive reviews by quoting the comments
\end{itemize}
\textbf{Capability \#18:}
\begin{itemize}[noitemsep,leftmargin=*]
   \item Cancel order \texttt{\tiny \{\{id\}\}}
\end{itemize}
\textbf{Capability \#19:}
\begin{itemize}[noitemsep,leftmargin=*]
   \item Change the page title of ``\texttt{\tiny \{\{old-heading\}\}}'' page on my site to ``\texttt{\tiny \{\{heading\}\}}''.
\end{itemize}
\textbf{Capability \#20:}
\begin{itemize}[noitemsep,leftmargin=*]
   \item Notify \texttt{\tiny \{\{name\}\}} in their most recent pending order with message ``\texttt{\tiny \{\{message\}\}}''
\end{itemize}
\textbf{Capability \#21:}
\begin{itemize}[noitemsep,leftmargin=*]
   \item Update order \#\texttt{\tiny \{\{order\}\}} with the \texttt{\tiny \{\{service\}\}} tracking number \texttt{\tiny \{\{tracking\}\}}
\end{itemize}
\textbf{Capability \#22:}
\begin{itemize}[noitemsep,leftmargin=*]
   \item Make all \texttt{\tiny \{\{product\}\}} as out of stock
\end{itemize}
\textbf{Capability \#23:}
\begin{itemize}[noitemsep,leftmargin=*]
   \item Modify the address of order \#\texttt{\tiny \{\{order\_id\}\}} to \texttt{\tiny \{\{address\}\}}
\end{itemize}
\textbf{Capability \#24:}
\begin{itemize}[noitemsep,leftmargin=*]
   \item Add new \texttt{\tiny \{\{option\}\}} \texttt{\tiny \{\{value\}\}} to \texttt{\tiny \{\{base\_setting\}\}} of \texttt{\tiny \{\{product\}\}}
\end{itemize}
\textbf{Capability \#25:}
\begin{itemize}[noitemsep,leftmargin=*]
   \item Lookup orders that are \texttt{\tiny \{\{status\}\}}
   \item Get the \texttt{\tiny \{\{attribute\}\}} of the \texttt{\tiny \{\{status\}\}} order
\end{itemize}
\textbf{Capability \#26:}
\begin{itemize}[noitemsep,leftmargin=*]
   \item Add a simple product named \texttt{\tiny \{\{product\}\}} with \texttt{\tiny \{\{stock\}\}} in stock, available in size \texttt{\tiny \{\{size\}\}} and color \texttt{\tiny \{\{color\}\}}, priced at \$\texttt{\tiny \{\{price\}\}}
\end{itemize}
\textbf{Capability \#27:}
\begin{itemize}[noitemsep,leftmargin=*]
   \item Draft a new marketing price rule for \texttt{\tiny \{\{topic\}\}} that offers \texttt{\tiny \{\{rule\}\}} for all customers
\end{itemize}
\textbf{Capability \#28:}
\begin{itemize}[noitemsep,leftmargin=*]
   \item Today is 3/15/2023, generate a \texttt{\tiny \{\{report\}\}} \texttt{\tiny \{\{time\_span\}\}}
   \item Create a \texttt{\tiny \{\{type\}\}} report from \texttt{\tiny \{\{start\_date\}\}} to \texttt{\tiny \{\{end\_date\}\}}
\end{itemize}
} & {
\textbf{Capability \#29:}
\begin{itemize}[noitemsep,leftmargin=*]
   \item We've received \texttt{\tiny \{\{quantity\}\}}, update the inventory.
\end{itemize}
\textbf{Capability \#30:}
\begin{itemize}[noitemsep,leftmargin=*]
   \item Approve the positive reviews to display in our store.
\end{itemize}
\textbf{Capability \#31:}
\begin{itemize}[noitemsep,leftmargin=*]
   \item Delete all \texttt{\tiny \{\{review\_type\}\}}
\end{itemize}
\textbf{Capability \#32:}
\begin{itemize}[noitemsep,leftmargin=*]
   \item Tell me the full address of all \texttt{\tiny \{\{airport\_type\}\}} that are within a driving distance of \texttt{\tiny \{\{radius\}\}} to \texttt{\tiny \{\{start\}\}}
\end{itemize}
\textbf{Capability \#33:}
\begin{itemize}[noitemsep,leftmargin=*]
   \item What is the \texttt{\tiny \{\{information\}\}} of \texttt{\tiny \{\{location\}\}}
   \item I will arrive \texttt{\tiny \{\{place\}\}} soon. Provide the name of a \texttt{\tiny \{\{target1\}\}} in the vicinity, if available. Then, tell me the \texttt{\tiny \{\{information\}\}} to \texttt{\tiny \{\{target2\}\}} from the hotel.
\end{itemize}
\textbf{Capability \#34:}
\begin{itemize}[noitemsep,leftmargin=*]
   \item What is the zip code of \texttt{\tiny \{\{place\}\}}?
\end{itemize}
\textbf{Capability \#35:}
\begin{itemize}[noitemsep,leftmargin=*]
   \item Given the following locations, \texttt{\tiny \{\{place\_list\}\}}, what would be the optimal route to travel through them all in order to minimize total travel time? Please note the journey begins at the first place listed.
\end{itemize}
\textbf{Capability \#36:}
\begin{itemize}[noitemsep,leftmargin=*]
   \item Which US states border \texttt{\tiny \{\{state\}\}}?
\end{itemize}
\textbf{Capability \#37:}
\begin{itemize}[noitemsep,leftmargin=*]
   \item Where is the nearest \texttt{\tiny \{\{places\}\}} to \texttt{\tiny \{\{start\}\}}, and what is the walking distance to it?
   \item Find the walkway to the closest \texttt{\tiny \{\{store\}\}} from \texttt{\tiny \{\{location\}\}}.
   \item How long does it take to walk from \texttt{\tiny \{\{start\}\}} to \texttt{\tiny \{\{end\}\}}?
   \item Tell me the closest \texttt{\tiny \{\{place1\}\}}(s) to \texttt{\tiny \{\{place2\}\}}
\end{itemize}
\textbf{Capability \#38:}
\begin{itemize}[noitemsep,leftmargin=*]
   \item From my stay at \texttt{\tiny \{\{hotel\}\}}, what's the estimated driving time to reach \texttt{\tiny \{\{place\}\}}?
   \item What is the minimum travel time by car from \texttt{\tiny \{\{location1\}\}} to \texttt{\tiny \{\{location2\}\}}?
   \item What is the duration required to first walk from \texttt{\tiny \{\{place\_A\}\}} to \texttt{\tiny \{\{place\_B\}\}}, and then drive to \texttt{\tiny \{\{place\_C\}\}}?
   \item Show me the walking distance from nearby hotels to \texttt{\tiny \{\{location\}\}} that take at most \texttt{\tiny \{\{n\}\}} minutes?
   \item What is the estimated driving time between \texttt{\tiny \{\{city1\}\}} and \texttt{\tiny \{\{city2\}\}}?
\end{itemize}
} \\
{
\textbf{Capability \#39:}
\begin{itemize}[noitemsep,leftmargin=*]
   \item From my stay at \texttt{\tiny \{\{hotel\}\}}, what's the estimated driving time to reach \texttt{\tiny \{\{place\}\}}?
   \item What is the estimated driving time between \texttt{\tiny \{\{city1\}\}} and \texttt{\tiny \{\{city2\}\}}?
   \item I am at CMU Pittsburgh, how long it takes to drive to the nearest \texttt{\tiny \{\{location\}\}}
   \item Check if the \texttt{\tiny \{\{place\}\}} in pittsburgh can be reached in one hour by car from \texttt{\tiny \{\{location\}\}}
\end{itemize}
\textbf{Capability \#40:}
\begin{itemize}[noitemsep,leftmargin=*]
   \item Find the \texttt{\tiny \{\{space\}\}} around \texttt{\tiny \{\{location\}\}}
   \item Find the walkway to the closest \texttt{\tiny \{\{store\}\}} from \texttt{\tiny \{\{location\}\}}.
   \item Tell me the closest \texttt{\tiny \{\{place1\}\}}(s) to \texttt{\tiny \{\{place2\}\}}
   \item Where is the nearest \texttt{\tiny \{\{location\}\}} from \texttt{\tiny \{\{location2\}\}} \texttt{\tiny \{\{condition\}\}}
\end{itemize}
\textbf{Capability \#41:}
\begin{itemize}[noitemsep,leftmargin=*]
   \item What is the \texttt{\tiny \{\{information\}\}} of \texttt{\tiny \{\{location\}\}}
   \item Tell me the coordinates of \texttt{\tiny \{\{location\}\}} in DD format
\end{itemize}
\textbf{Capability \#42:}
\begin{itemize}[noitemsep,leftmargin=*]
   \item How much time does it take from Pittsburgh to Philadelphia by car?
\end{itemize}
\textbf{Capability \#43:}
\begin{itemize}[noitemsep,leftmargin=*]
   \item Show the route from SCS CMU in Pittsburgh to the location where the Declaration of Independence and Constitution were signed
\end{itemize}
\textbf{Capability \#44:}
\begin{itemize}[noitemsep,leftmargin=*]
   \item Pull up the description page of \texttt{\tiny \{\{location\}\}} on Map
   \item What is the \texttt{\tiny \{\{information\}\}} of \texttt{\tiny \{\{location\}\}}
\end{itemize}
\textbf{Capability \#45:}
\begin{itemize}[noitemsep,leftmargin=*]
   \item I am arriving at Carnegie Mellon University. Find the nearby US Citizenship and Immigration Services and the walking distance to the nearest Social Security Administration from US Citizenship and Immigration Services
\end{itemize}
\textbf{Capability \#46:}
\begin{itemize}[noitemsep,leftmargin=*]
   \item I am arriving at Pittsburgh Airport. Show me the name of a Hyatt hotel if there is any nearby. Tell me the names of supermarkets that are within 15mins driving from the hotel
\end{itemize}
\textbf{Capability \#47:}
\begin{itemize}[noitemsep,leftmargin=*]
   \item Measure distance between \texttt{\tiny \{\{location/address\_1\}\}} and \texttt{\tiny \{\{location/address\_2\}\}} by walking
   \item Get directions from \texttt{\tiny \{\{location/address\_1\}\}} to \texttt{\tiny \{\{location/address\_2\}\}} using \texttt{\tiny \{\{transportation\}\}} options.
\end{itemize}
\textbf{Capability \#48:}
\begin{itemize}[noitemsep,leftmargin=*]
   \item List out reviewers, if exist, who mention about \texttt{\tiny \{\{description\}\}}
\end{itemize}
\textbf{Capability \#49:}
\begin{itemize}[noitemsep,leftmargin=*]
   \item Today is 6/12/2023. Tell me how many fulfilled orders I have \texttt{\tiny \{\{period\}\}}, and the total amount of money I spent.
\end{itemize}
} & {
\textbf{Capability \#50:}
\begin{itemize}[noitemsep,leftmargin=*]
   \item Tell me the status of my latest order and when will it arrive
\end{itemize}
\textbf{Capability \#51:}
\begin{itemize}[noitemsep,leftmargin=*]
   \item What is the date when I made my first purchase on this site?
\end{itemize}
\textbf{Capability \#52:}
\begin{itemize}[noitemsep,leftmargin=*]
   \item I have jaw bruxism problem, show me something that could alleviate the problem.
\end{itemize}
\textbf{Capability \#53:}
\begin{itemize}[noitemsep,leftmargin=*]
   \item What is the price range for products from \texttt{\tiny \{\{brand\}\}}?
   \item What is the price range of \texttt{\tiny \{\{product\}\}} in the One Stop Market?
\end{itemize}
\textbf{Capability \#54:}
\begin{itemize}[noitemsep,leftmargin=*]
   \item How much I spent on \texttt{\tiny \{\{category\}\}} shopping during \texttt{\tiny \{\{time\}\}}
\end{itemize}
\textbf{Capability \#55:}
\begin{itemize}[noitemsep,leftmargin=*]
   \item What is the \texttt{\tiny \{\{option\}\}} configuration of the \texttt{\tiny \{\{product\}\}} I bought \texttt{\tiny \{\{time\}\}}
   \item I previously ordered some \texttt{\tiny \{\{product\}\}} \texttt{\tiny \{\{time\}\}} and later cancelled. Can you reorder it for me?
\end{itemize}
\textbf{Capability \#56:}
\begin{itemize}[noitemsep,leftmargin=*]
   \item I have a lot of Nintendo Switch game cards now, help me find the best storage option to fit all \texttt{\tiny \{\{num\}\}} cards
\end{itemize}
\textbf{Capability \#57:}
\begin{itemize}[noitemsep,leftmargin=*]
   \item What are the main criticisms of this product? Please extract the relevant sentences.
\end{itemize}
\textbf{Capability \#58:}
\begin{itemize}[noitemsep,leftmargin=*]
   \item What do customers say about \texttt{\tiny \{\{product\_type\}\}} from \texttt{\tiny \{\{manufature\}\}}
\end{itemize}
\textbf{Capability \#59:}
\begin{itemize}[noitemsep,leftmargin=*]
   \item Buy the best rating product from ``\texttt{\tiny \{\{category\}\}}'' category with at least 5 reviews and the product is least expensive
   \item I am doing a market survey for one stop market, show me the most expensive product from \texttt{\tiny \{\{product\_category\}\}} category
   \item Buy the highest rated product from the \texttt{\tiny \{\{product\_category\}\}} category within a budget \texttt{\tiny \{\{dollar\_value\}\}}.
\end{itemize}
\textbf{Capability \#60:}
\begin{itemize}[noitemsep,leftmargin=*]
   \item Search for ``\texttt{\tiny \{\{keyword\}\}}''
\end{itemize}
\textbf{Capability \#61:}
\begin{itemize}[noitemsep,leftmargin=*]
   \item List the full product names of slide slippers from Nike and tell me the price range of the available products
\end{itemize}
\textbf{Capability \#62:}
\begin{itemize}[noitemsep,leftmargin=*]
   \item Look up the most recent models of XBox controllers released between 2020-2021?
\end{itemize}
\textbf{Capability \#63:}
\begin{itemize}[noitemsep,leftmargin=*]
   \item Show the least expensive \texttt{\tiny \{\{product\}\}} with a minimum storage capacity of \texttt{\tiny \{\{min\_storage\}\}}.
\end{itemize}
\textbf{Capability \#64:}
\begin{itemize}[noitemsep,leftmargin=*]
   \item Show the most recent \texttt{\tiny \{\{status\}\}} order
   \item Get the order number of my most recent \texttt{\tiny \{\{status\}\}} order 
\end{itemize}
} & {
\textbf{Capability \#65:}
\begin{itemize}[noitemsep,leftmargin=*]
   \item Which number to call for the customer service?
\end{itemize}
\textbf{Capability \#66:}
\begin{itemize}[noitemsep,leftmargin=*]
   \item How much refund I should expect from my order canlled in \texttt{\tiny \{\{time\}\}}? I only kept the AC-DC Adapter and the shop told me that I cannot get the shipping fee back
\end{itemize}
\textbf{Capability \#67:}
\begin{itemize}[noitemsep,leftmargin=*]
   \item Show me the ``\texttt{\tiny \{\{product\}\}}'' listings by \texttt{\tiny \{\{sorting\_order\}\}}.
\end{itemize}
\textbf{Capability \#68:}
\begin{itemize}[noitemsep,leftmargin=*]
   \item How much did I spend on shopping at One Stop Market \texttt{\tiny \{\{time\}\}}? They gave me a 20\% discount on the total amount for orders exceeding \$200 in cash
\end{itemize}
\textbf{Capability \#69:}
\begin{itemize}[noitemsep,leftmargin=*]
   \item Tell me when I last ordered my \texttt{\tiny \{\{description\}\}}?
\end{itemize}
\textbf{Capability \#70:}
\begin{itemize}[noitemsep,leftmargin=*]
   \item List products from \texttt{\tiny \{\{product\_category\}\}} category by \texttt{\tiny \{\{order\}\}} price
   \item Show me products under \$\texttt{\tiny \{\{price\}\}} in ``\texttt{\tiny \{\{product\_category\}\}}'' category
\end{itemize}
\textbf{Capability \#71:}
\begin{itemize}[noitemsep,leftmargin=*]
   \item Show me the \texttt{\tiny \{\{info\}\}} for order number \texttt{\tiny \{\{order\_number\}\}}.
\end{itemize}
\textbf{Capability \#72:}
\begin{itemize}[noitemsep,leftmargin=*]
   \item find discounted items.
\end{itemize}
\textbf{Capability \#73:}
\begin{itemize}[noitemsep,leftmargin=*]
   \item Summarize customer reviews for \texttt{\tiny \{\{product\}\}}.
\end{itemize}
\textbf{Capability \#74:}
\begin{itemize}[noitemsep,leftmargin=*]
   \item List the customer names who thinks EYZUTAK phone cases are of good looking
   \item Who gave \texttt{\tiny \{\{stars\}\}} for phone cases from EYZUTAK
\end{itemize}
\textbf{Capability \#75:}
\begin{itemize}[noitemsep,leftmargin=*]
   \item What is the rating of \texttt{\tiny \{\{product\}\}}
\end{itemize}
\textbf{Capability \#76:}
\begin{itemize}[noitemsep,leftmargin=*]
   \item Add the product with the lowest per unit price from my open tabs to the shopping cart
\end{itemize}
\textbf{Capability \#77:}
\begin{itemize}[noitemsep,leftmargin=*]
   \item Add \texttt{\tiny \{\{product\}\}} to my wish list
   \item Add this product to my wishlist
   \item Add a \texttt{\tiny \{\{product\}\}} to my wish list.
\end{itemize}
\textbf{Capability \#78:}
\begin{itemize}[noitemsep,leftmargin=*]
   \item Subscribe to the newsletter of OneStopMarket
\end{itemize}
\textbf{Capability \#79:}
\begin{itemize}[noitemsep,leftmargin=*]
   \item I recently moved, my address is \texttt{\tiny \{\{address\}\}}, update my information on OneStopShopping accordingly
   \item Change the delivery address for my most recent order to \texttt{\tiny \{\{address\}\}}.
\end{itemize}
\textbf{Capability \#80:}
\begin{itemize}[noitemsep,leftmargin=*]
   \item Rate my recent purchase of \texttt{\tiny \{\{product\}\}} with \texttt{\tiny \{\{num\_star\}\}} stars, using my nickname \texttt{\tiny \{\{nickname\}\}}?
\end{itemize}
} \\
{
\textbf{Capability \#81:}
\begin{itemize}[noitemsep,leftmargin=*]
   \item Fill the ``contact us'' form in the site for a refund on the \texttt{\tiny \{\{product\}\}} I bought, stating that it broke after just three days of use. Also, ensure to include the order number \#\texttt{\tiny \{\{order\_id\}\}} and the product SKU. Don't submit yet, I will check.
   \item Draft a refund message via their ``contact us'' form for the \texttt{\tiny \{\{product\}\}} I bought \texttt{\tiny \{\{time\}\}}. It broke after three days of use. The shop requires the order id, the reason and the amount to refund in the message. Don't submit yet
\end{itemize}
\textbf{Capability \#82:}
\begin{itemize}[noitemsep,leftmargin=*]
   \item Draft an email to the shop owner via their contact us function for a coupon as \texttt{\tiny \{\{reason\}\}}
\end{itemize}
\textbf{Capability \#83:}
\begin{itemize}[noitemsep,leftmargin=*]
   \item Tell me the count of comments that have received more downvotes than upvotes for the user who made the latest post on the \texttt{\tiny \{\{forum\}\}} forum.
\end{itemize}
\textbf{Capability \#84:}
\begin{itemize}[noitemsep,leftmargin=*]
   \item Among the top \texttt{\tiny \{\{number\}\}} post in ``\texttt{\tiny \{\{subreddit\}\}}'' forum, \texttt{\tiny \{\{description\}\}}
\end{itemize}
\textbf{Capability \#85:}
\begin{itemize}[noitemsep,leftmargin=*]
   \item Change my reddit bio to ``\texttt{\tiny \{\{content\}\}}''
\end{itemize}
\textbf{Capability \#86:}
\begin{itemize}[noitemsep,leftmargin=*]
   \item Reply to \texttt{\tiny \{\{position\_description\}\}} with my comment ``\texttt{\tiny \{\{content\_description\}\}}''
\end{itemize}
\textbf{Capability \#87:}
\begin{itemize}[noitemsep,leftmargin=*]
   \item Create a new forum named \texttt{\tiny \{\{name\}\}}, with a description of \texttt{\tiny \{\{description\}\}}, and include \texttt{\tiny \{\{sidebar\_list\}\}} in the sidebar?
\end{itemize}
\textbf{Capability \#88:}
\begin{itemize}[noitemsep,leftmargin=*]
   \item Open the thread of a trending post on the forum ``\texttt{\tiny \{\{subreddit\}\}}'' and subscribe.
   \item Upvote the newest post in \texttt{\tiny \{\{subreddit\}\}} subreddit
\end{itemize}
\textbf{Capability \#89:}
\begin{itemize}[noitemsep,leftmargin=*]
   \item Create a discussion post about ``\texttt{\tiny \{\{topic\}\}}'' in a relevant subreddit and ask users for their opinions with the simple prompt, ``your opinion''
   \item Find a subreddit focused on topics related to \texttt{\tiny \{\{topic\}\}}, and post my question, ``\texttt{\tiny \{\{question\}\}}'' there
   \item Post my question, ``\texttt{\tiny \{\{question\}\}}\ in a subreddit where I'm likely to get an answer
\end{itemize}
\textbf{Capability \#90:}
\begin{itemize}[noitemsep,leftmargin=*]
   \item Post a review of my recent reading ``\texttt{\tiny \{\{book\}\}}'' in the r/books with my comment ``\texttt{\tiny \{\{content\}\}}''.
\end{itemize}
\textbf{Capability \#91:}
\begin{itemize}[noitemsep,leftmargin=*]
   \item Re-post the image of \texttt{\tiny \{\{content\}\}} in this page to \texttt{\tiny \{\{subreddit\}\}} subreddit and note ``from /f/pics''
\end{itemize}
\textbf{Capability \#92:}
\begin{itemize}[noitemsep,leftmargin=*]
   \item Ask for advice about \texttt{\tiny \{\{issue\}\}} in a subreddit for relations
\end{itemize}
} & {
\textbf{Capability \#93:}
\begin{itemize}[noitemsep,leftmargin=*]
   \item Post in the most appropriate subreddit and ask for recommendations for \texttt{\tiny \{\{category\}\}} products within a budget of \texttt{\tiny \{\{price\}\}}
   \item Ask for product recommendations for \texttt{\tiny \{\{category\}\}} within a budget of \texttt{\tiny \{\{price\}\}} in \texttt{\tiny \{\{subreddit\}\}}
\end{itemize}
\textbf{Capability \#94:}
\begin{itemize}[noitemsep,leftmargin=*]
   \item Post a notice on a virtual meetup for \texttt{\tiny \{\{interest\}\}} enthusiasts on \texttt{\tiny \{\{date\}\}} in the \texttt{\tiny \{\{subreddit\}\}} subreddit
\end{itemize}
\textbf{Capability \#95:}
\begin{itemize}[noitemsep,leftmargin=*]
   \item Post in \texttt{\tiny \{\{subreddit\}\}} subreddit about what could diffusion model help the correpong field.
\end{itemize}
\textbf{Capability \#96:}
\begin{itemize}[noitemsep,leftmargin=*]
   \item Thumbs down the top \texttt{\tiny \{\{k\}\}} post ever in \texttt{\tiny \{\{subreddit\}\}}.
\end{itemize}
\textbf{Capability \#97:}
\begin{itemize}[noitemsep,leftmargin=*]
   \item Like all submissions created by \texttt{\tiny \{\{user\}\}} in subreddit \texttt{\tiny \{\{subreddit\}\}}
   \item DisLike all submissions created by \texttt{\tiny \{\{user\}\}} in subreddit \texttt{\tiny \{\{subreddit\}\}}
\end{itemize}
\textbf{Capability \#98:}
\begin{itemize}[noitemsep,leftmargin=*]
   \item Edit my post on \texttt{\tiny \{\{post\}\}} by adding a line to the body that says ``\texttt{\tiny \{\{content\}\}}''
\end{itemize}
\textbf{Capability \#99:}
\begin{itemize}[noitemsep,leftmargin=*]
   \item Check out my todos
\end{itemize}
\textbf{Capability \#100:}
\begin{itemize}[noitemsep,leftmargin=*]
   \item Check out the most recent open issues
\end{itemize}
\textbf{Capability \#101:}
\begin{itemize}[noitemsep,leftmargin=*]
   \item Checkout merge requests assigned to me
   \item Checkout merge requests requiring my review
\end{itemize}
\textbf{Capability \#102:}
\begin{itemize}[noitemsep,leftmargin=*]
   \item Tell me the full names of the repositories where I made contributions and they got \texttt{\tiny \{\{description\}\}} stars?
\end{itemize}
\textbf{Capability \#103:}
\begin{itemize}[noitemsep,leftmargin=*]
   \item Open my latest created issue that has \texttt{\tiny \{\{keyword\}\}} in its title to check if it is closed
   \item Open my latest updated issue that has keyword ``\texttt{\tiny \{\{keyword\}\}}'' in its title to check if it is closed
\end{itemize}
\textbf{Capability \#104:}
\begin{itemize}[noitemsep,leftmargin=*]
   \item See all public projects
\end{itemize}
\textbf{Capability \#105:}
\begin{itemize}[noitemsep,leftmargin=*]
   \item Get me my RSS feed token
\end{itemize}
\textbf{Capability \#106:}
\begin{itemize}[noitemsep,leftmargin=*]
   \item Show me the command to clone \texttt{\tiny \{\{repo\}\}} with SSH.
\end{itemize}
\textbf{Capability \#107:}
\begin{itemize}[noitemsep,leftmargin=*]
   \item List all opened issues \texttt{\tiny \{\{description\}\}}
\end{itemize}
\textbf{Capability \#108:}
\begin{itemize}[noitemsep,leftmargin=*]
   \item Who else have access to my repo \texttt{\tiny \{\{repo\}\}}, show me their usernames
\end{itemize}
} & {
\textbf{Capability \#109:}
\begin{itemize}[noitemsep,leftmargin=*]
   \item Post ``\texttt{\tiny \{\{content\}\}}'' for the merge request related to \texttt{\tiny \{\{mr\}\}} in \texttt{\tiny \{\{repo\}\}} project
\end{itemize}
\textbf{Capability \#110:}
\begin{itemize}[noitemsep,leftmargin=*]
   \item Fork \texttt{\tiny \{\{repo\}\}}.
\end{itemize}
\textbf{Capability \#111:}
\begin{itemize}[noitemsep,leftmargin=*]
   \item Make the LICENSE of \texttt{\tiny \{\{repo\}\}} to MIT license.
\end{itemize}
\textbf{Capability \#112:}
\begin{itemize}[noitemsep,leftmargin=*]
   \item Go to the merge request on \texttt{\tiny \{\{topic\}\}} I have to review, find if the author of the merge request responded at the end, and reply ``Thank you'' if he did. Otherwise remind him with a simple @.
\end{itemize}
\textbf{Capability \#113:}
\begin{itemize}[noitemsep,leftmargin=*]
   \item Set my gitlab status as \texttt{\tiny \{\{status\}\}}.
\end{itemize}
\textbf{Capability \#114:}
\begin{itemize}[noitemsep,leftmargin=*]
   \item Update the project site's title to ``\texttt{\tiny \{\{title\}\}}''
\end{itemize}
\textbf{Capability \#115:}
\begin{itemize}[noitemsep,leftmargin=*]
   \item set the homepage URL on my GitLab profile to \texttt{\tiny \{\{url\}\}}
\end{itemize}
\textbf{Capability \#116:}
\begin{itemize}[noitemsep,leftmargin=*]
   \item Set up a new, empty repository with the name \texttt{\tiny \{\{project\_name\}\}}?
   \item Create a private \texttt{\tiny \{\{template\}\}} repository called ``\texttt{\tiny \{\{project\_name\}\}}'' using the right template to speed up development.
\end{itemize}
\textbf{Capability \#117:}
\begin{itemize}[noitemsep,leftmargin=*]
   \item Invite \newline \texttt{\tiny \{\{collaborator\_account\_list\}\}} as collaborator to \texttt{\tiny \{\{repo\}\}} repo
   \item Add the following users to repo \texttt{\tiny \{\{repo\}\}} as \texttt{\tiny \{\{role\}\}}: \texttt{\tiny \{\{user\_list\}\}}
\end{itemize}
\textbf{Capability \#118:}
\begin{itemize}[noitemsep,leftmargin=*]
   \item \texttt{\tiny \{\{name\}\}} wants to check my dotfile configurations. Please invite him to the repo as a guest.
\end{itemize}
\textbf{Capability \#119:}
\begin{itemize}[noitemsep,leftmargin=*]
   \item Star the top \texttt{\tiny \{\{number\}\}} most stared repos in Gitlab
\end{itemize}
\textbf{Capability \#120:}
\begin{itemize}[noitemsep,leftmargin=*]
   \item Follow \texttt{\tiny \{\{account\_list\}\}} on Gitlab
\end{itemize}
\textbf{Capability \#121:}
\begin{itemize}[noitemsep,leftmargin=*]
   \item Create a milestone for the upcoming \texttt{\tiny \{\{event\}\}} starting on \texttt{\tiny \{\{start\_date\}\}} and ending on \texttt{\tiny \{\{end\_date\}\}}
\end{itemize}
\textbf{Capability \#122:}
\begin{itemize}[noitemsep,leftmargin=*]
   \item Create an issue \texttt{\tiny \{\{issue\}\}} in \texttt{\tiny \{\{repo\}\}}.
   \item Assign the issue regarding \texttt{\tiny \{\{issue\}\}} in \texttt{\tiny \{\{repo\}\}} to \texttt{\tiny \{\{account\}\}}.
   \item Create an issue in \texttt{\tiny \{\{repo\}\}} repo with title ``\texttt{\tiny \{\{issue\}\}}''. Assign the issue to \texttt{\tiny \{\{account\}\}}. Set due date to be \texttt{\tiny \{\{due\}\}}
\end{itemize}
} \\
{
\textbf{Capability \#123:}
\begin{itemize}[noitemsep,leftmargin=*]
   \item Submit a merge request for \texttt{\tiny \{\{source\_branch\}\}} branch to be merged into \texttt{\tiny \{\{target\_branch\}\}} branch, assign \texttt{\tiny \{\{reviewer\}\}} as the reviewer
\end{itemize}
\textbf{Capability \#124:}
\begin{itemize}[noitemsep,leftmargin=*]
   \item Open a new issue to discuss the implementation of \texttt{\tiny \{\{feature\}\}}
\end{itemize}
\textbf{Capability \#125:}
\begin{itemize}[noitemsep,leftmargin=*]
   \item Start a private project \texttt{\tiny \{\{project\_name\}\}} with \texttt{\tiny \{\{template\}\}} template and add \texttt{\tiny \{\{account\_list\}\}} as members
\end{itemize}
\textbf{Capability \#126:}
\begin{itemize}[noitemsep,leftmargin=*]
   \item How many commits did \texttt{\tiny \{\{user\}\}} make on \texttt{\tiny \{\{date\}\}} in total?
   \item Tell me who has made the most contributions, in terms of number of commits, to the \texttt{\tiny \{\{repo\}\}} project
   \item Tell me the \texttt{\tiny \{\{attribute\}\}} of the contributor who has the most commits to branch \texttt{\tiny \{\{branch\_name\}\}}
   \item List the \texttt{\tiny \{\{attribute\}\}} of the top 3 contributors to \texttt{\tiny \{\{repo\}\}} repo, ranked by the number of commits?
\end{itemize}
\textbf{Capability \#127:}
\begin{itemize}[noitemsep,leftmargin=*]
   \item create a new group ``\texttt{\tiny \{\{name\}\}}'' with members \texttt{\tiny \{\{members\}\}}
\end{itemize}
\textbf{Capability \#128:}
\begin{itemize}[noitemsep,leftmargin=*]
   \item Tell me the distance to drive from Carnegie Mellon University to the top computer science school in massachusetts
\end{itemize}
\textbf{Capability \#129:}
\begin{itemize}[noitemsep,leftmargin=*]
   \item What's the closest national park to \texttt{\tiny \{\{city\}\}}? How long does it take to bike there?
\end{itemize}
\textbf{Capability \#130:}
\begin{itemize}[noitemsep,leftmargin=*]
   \item Find the page of \texttt{\tiny \{\{description\}\}} on the map.
\end{itemize}
\textbf{Capability \#131:}
\begin{itemize}[noitemsep,leftmargin=*]
   \item Show me the way from \texttt{\tiny \{\{location\}\}} to the home stadium of \texttt{\tiny \{\{sport\_team\}\}} \texttt{\tiny \{\{time\}\}}
\end{itemize}
\textbf{Capability \#132:}
\begin{itemize}[noitemsep,leftmargin=*]
   \item Find a GitLab repository related to \texttt{\tiny \{\{topic\}\}} and make a Reddit post linking to it in a relevant subreddit
   \item create a repository named \texttt{\tiny \{\{name\}\}} that includes a README file with the links to the most active \texttt{\tiny \{\{num\}\}} DIY ideas on DIY subreddit?
   \item Make a folder named \texttt{\tiny \{\{directory\}\}} on the \texttt{\tiny \{\{gitlab\_repo\}\}} repo and include a file called urls.txt that consists of the links to the 5 most recent posts from \texttt{\tiny \{\{subreddit\}\}}.
\end{itemize}
\textbf{Capability \#133:}
\begin{itemize}[noitemsep,leftmargin=*]
   \item Promote \texttt{\tiny \{\{repo\}\}} to subreddit \texttt{\tiny \{\{subreddit\}\}} with the description from the repo itself.
\end{itemize}
\textbf{Capability \#134:}
\begin{itemize}[noitemsep,leftmargin=*]
   \item Create a repo named \texttt{\tiny \{\{name\}\}} with \texttt{\tiny \{\{topics\}\}} in a README file
\end{itemize}
} & {
\textbf{Capability \#135:}
\begin{itemize}[noitemsep,leftmargin=*]
   \item Gather the titles of \texttt{\tiny \{\{product\}\}} reviews with \texttt{\tiny \{\{rating\}\}} rating from OneStopShop, and post them in the games subreddit under the title ``real user feedback on \texttt{\tiny \{\{product\}\}}''
\end{itemize}
\textbf{Capability \#136:}
\begin{itemize}[noitemsep,leftmargin=*]
   \item Show me the route and driving time from \texttt{\tiny \{\{city1\}\}} to \texttt{\tiny \{\{city2\}\}}
\end{itemize}
} & {

} \\
\end{longtable}
}

\raggedbottom
\newpage
\section{Prompts}
\label{sec:appendix_prompts}

We provide the prompts used to generate novel out-of-domain objectives, urls, web pages, and solution trajectories.
\newline

\textbf{Generate Novel Synthetic Objectives and Websites:}
\newline

\noindent{\scriptsize\texttt{Here are a few example objectives (tasks) a user might be asked to perform on a webpage. Closely following these example objectives, generate a potential objective a user might want to perform on another American website that is similar to the examples. (in terms of reasoning required, requiring navigating to multiple pages or taking multiple steps to solve, etc.) The new objective should not be on a website that is the same or is similar to any of the example objective's websites/domains, it should be a completely different website. Ensure the objective has a definitive, objective answer, and not a subjective answer. Return just the objective and a domain name (no path in the URL, just the hostname) of the website (in the same OBJECTIVE:/URL: format) and nothing else.
\newline\newline
OBJECTIVE: \{...\}\newline
URL: \{...\}\newline\newline
\{...other examples\}
}}
\newline

\textbf{Generate Plan for Hypothetical Synthetic Solution Trajectory:}
\newline

\noindent{\scriptsize\texttt{OBJECTIVE: \{...\}\newline URL: \{...\}\newline\newline Here is an objective a user can perform on the webpage. The user may need to perform multiple actions / steps (clicking, typing, scrolling, storing/remembering information, or recalling stored information) in order to solve the objective. Assuming the user is starting with a web browser that is already loaded with the website, output the required / necessary steps the user must take on the page to solve the objective, one step per line. Each step MUST involve either clicking, scrolling, typing, or stopping (when the objective is complete). DO NOT output steps that don't involve one of these actions. If a step does not involve clicking, scrolling, typing, or stopping, such as remembering/recalling/calculating information, combine it instead with the next step in the sequence that does. Return nothing else other than the necessary steps, no bullets and no numbered lists.
}}
\newline

\textbf{Generate Hypothetical URL for Random Step in Synthetic Trajectory:}
\newline

\noindent{\scriptsize\texttt{OBJECTIVE: \{...\}\newline WEBSITE: \{...\}\newline STEPS:\newline 1. \{...\}\newline2. \{...\}\newline\{...other steps\}\newline\newline Here is an objective a user can perform on a website starting from the homepage and some steps a user may take to solve the objective. Output a realistic and valid URL (don't use placeholders like '123', 'example', 'acme', etc.) for what page a user would be on after they perform Step \#\{...\}. Return just the URL and nothing else.
}}
\raggedbottom
\newpage

\textbf{Generate Hypothetical Previous and Next Action for Random Step in Synthetic Trajectory:}
\newline

\noindent{\scriptsize\texttt{Here are 2 example objectives a user might be asked to perform on a URL / webpage (provided in accessibility tree format). The goal is to perform a series of incremental actions that can complete the objective. The previous action that was taken and the next action a user should take towards completing the objective along with a "Let's think step-by-step." explanation is also provided for the 2 examples. All actions possible for the user are:\newline \newline \{...\}\newline \newline The action should always be placed inside \textasciigrave\textasciigrave\textasciigrave\textasciigrave\textasciigrave\textasciigrave. For example, "In summary, the next action I will perform is \textasciigrave\textasciigrave\textasciigrave click [1234]\textasciigrave\textasciigrave\textasciigrave".
\newline \newline Example 1:\newline \newline
OBJECTIVE: \{...\}\newline
URL: \{...\}\newline
WEBPAGE: \{...\}\newline
PREVIOUS ACTION: \{...\}\newline
NEXT ACTION: \{...\}\newline \newline
Example 2:\newline \newline
\{...other example\}\newline \newline
Following the structure of these 2 examples closely, for the objective and URL below, generate a realistic full-length webpage accessibility tree, realistic previous action, and realistic next action that a user needs to perform on the webpage in order to complete Step \#\{...\} of the OVERALL PLAN towards the objective. Provide the actions and webpage in the same format (WEBPAGE:/PREVIOUS ACTION:/NEXT ACTION:). Ensure the next action is Step \#\{...\}, the next action begins with "Let's think step-by-step." and ends with "In summary, the next action I will perform is \textasciigrave\textasciigrave\textasciigrave...\textasciigrave\textasciigrave\textasciigrave", and the [id] for any actions is an ID number not a string. Do not mention or reference the OVERALL PLAN or Step \#\{...\} directly in the output. Return nothing else other than the two actions and the webpage.\newline \newline 
OBJECTIVE: \{...\}\newline 
URL: \{...\}\newline 
OVERALL PLAN:\newline 1. \{...\}\newline2. \{...\}\newline\{...other steps\}\newline
CURRENT STEP: \{...\}
}}\newline

\raggedbottom\newpage
\textbf{Generate Hypothetical Web Page for Random Step in Synthetic Trajectory:}
\newline

\noindent{\scriptsize\texttt{Here are 2 example objectives a user might be asked to perform on a URL / webpage (provided in accessibility tree format). The goal is to perform a series of incremental actions that can complete the objective. The previous action that was taken and the next action a user should take towards completing the objective along with a "Let's think step-by-step." explanation is also provided for the 2 examples. All actions possible for the user are:\newline \newline \{...\}\newline \newline The action should always be placed inside \textasciigrave\textasciigrave\textasciigrave\textasciigrave\textasciigrave\textasciigrave. For example, "In summary, the next action I will perform is \textasciigrave\textasciigrave\textasciigrave click [1234]\textasciigrave\textasciigrave\textasciigrave".
\newline \newline
Example 1:\newline \newline
OBJECTIVE: \{...\}\newline
URL: \{...\}\newline
WEBPAGE: \{...\}\newline
PREVIOUS ACTION: \{...\}\newline
NEXT ACTION: \{...\}\newline \newline
Example 2:\newline \newline
\{...other example\}\newline \newline
Following the structure of these 2 examples closely, for the objective, URL, previous action, and next action below, generate a realistic full-length webpage accessibility tree (don't use placeholders like '123', 'example', 'acme', etc.). Ensure the page is in English and is structured such that performing the next action described would realistically complete or make incremental progress towards completing the objective. Provide the webpage in the same format (WEBPAGE:) and return nothing else other than the webpage.\newline\newline 
OBJECTIVE: \{...\}\newline
URL: \{...\}\newline
PREVIOUS ACTION: \{...\}\newline
NEXT ACTION: \{...\}
}}
\raggedbottom\newpage
\section{Resources}
\label{sec:appendix_resources}

\noindent We provide links and citations to resources used in this paper which provide license information, documentation, and their intended use. Our usage follows the intended usage of all resources.

~\\ \noindent We utilize the following models:
\begin{itemize}
    \small
    \item GPT-4 \citep{gpt4}
    \item Qwen-1.5-72B-Chat \citep{qwen}
    \item \texttt{\small sentence-transformers/all-distilroberta-v1} \citep{distilroberta,roberta}
    \item \texttt{\small sentence-transformers/all-mpnet-base-v2} \citep{Song:20mpnet}
\end{itemize}

~\\ \noindent We utilize the following datasets:
\begin{itemize}
    \small
    \item WebArena Benchmark \citep{webarena}
\end{itemize}

~\\ \noindent We utilize the following software:
\begin{itemize}
    \small
    \item DataDreamer \citep{datadreamer}
    \item PyTorch and PyTorch FSDP 
    \citep{pytorch,fsdp}
    \item QLora \citep{qlora}
    \item Transformers \citep{transformers}
    \item Sentence-Transformers \citep{sentencetransformers}
    \item SymbolicAI \citep{symbolicai}
    \item fastdtw \citep{fastdtw:software, Stan:04fastdtw}
\end{itemize}

~\\ \noindent We estimate the total compute budget and detail computing infrastructure used to run the computational experiments found in this paper below:
\begin{itemize}
    \small
    \item 4x NVIDIA RTX A6000 / 300GB RAM / 50x CPU -- 900 hours
\end{itemize}

\end{document}